\pdfoutput=1

\documentclass[11pt]{article}

\usepackage[]{EACL2023}

\usepackage{times}
\usepackage{latexsym}

\usepackage{amsmath}
\usepackage{booktabs}
\usepackage{graphicx}
\usepackage{subcaption}

\usepackage[T1]{fontenc}

\usepackage[utf8]{inputenc}

\usepackage{microtype}

\newcommand{\lookahead}{\textsc{Beam+Lookahead}}
\newcommand{\ranking}{\textsc{Beam+Ranking}}

\title{Faithfulness-Aware Decoding Strategies for Abstractive Summarization}

\author{David Wan$^1$\thanks{$^\ast$Work conducted during an internship at Amazon.} \,\, Mengwen Liu$^2$ \,\, Kathleen McKeown$^{3}$ \,\, \textbf{Markus Dreyer$^2$\thanks{$\dagger$Corresponding authors.} \,\, Mohit Bansal$^{1,2}$$^\dagger$} \\
$^1$UNC Chapel Hill \quad $^2$Amazon Alexa AI \quad $^3$AWS AI Labs \\
\texttt{\{davidwan,mbansal\}@cs.unc.edu} \\
\texttt{\{mengwliu,mckeownk,mddreyer,mobansal\}@amazon.com}
}

\begin{document}
\maketitle

\begin{abstract}
Despite significant progress in understanding and improving faithfulness in abstractive summarization, the question of how decoding strategies affect faithfulness is less studied. We present a systematic study of the effect of generation techniques such as beam search and nucleus sampling on faithfulness in abstractive summarization. We find a consistent trend where beam search with large beam sizes produces the most faithful summaries while nucleus sampling generates the least faithful ones. We propose two faithfulness-aware generation methods to further improve faithfulness over current generation techniques: (1) ranking candidates generated by beam search using automatic faithfulness metrics and (2) incorporating lookahead heuristics that produce a faithfulness score on the future summary. We show that both generation methods significantly improve faithfulness across two datasets as evaluated by four automatic faithfulness metrics and human evaluation. To reduce computational cost, we demonstrate a simple distillation approach that allows the model to generate faithful summaries with just greedy decoding.\footnote{Our code is publicly available at \url{https://github.com/amazon-science/faithful-summarization-generation}.}
\end{abstract}

\begin{figure*}[!ht]
    \centering
    \begin{subfigure}{.5\textwidth}
    \centering
    \includegraphics[width=\linewidth]{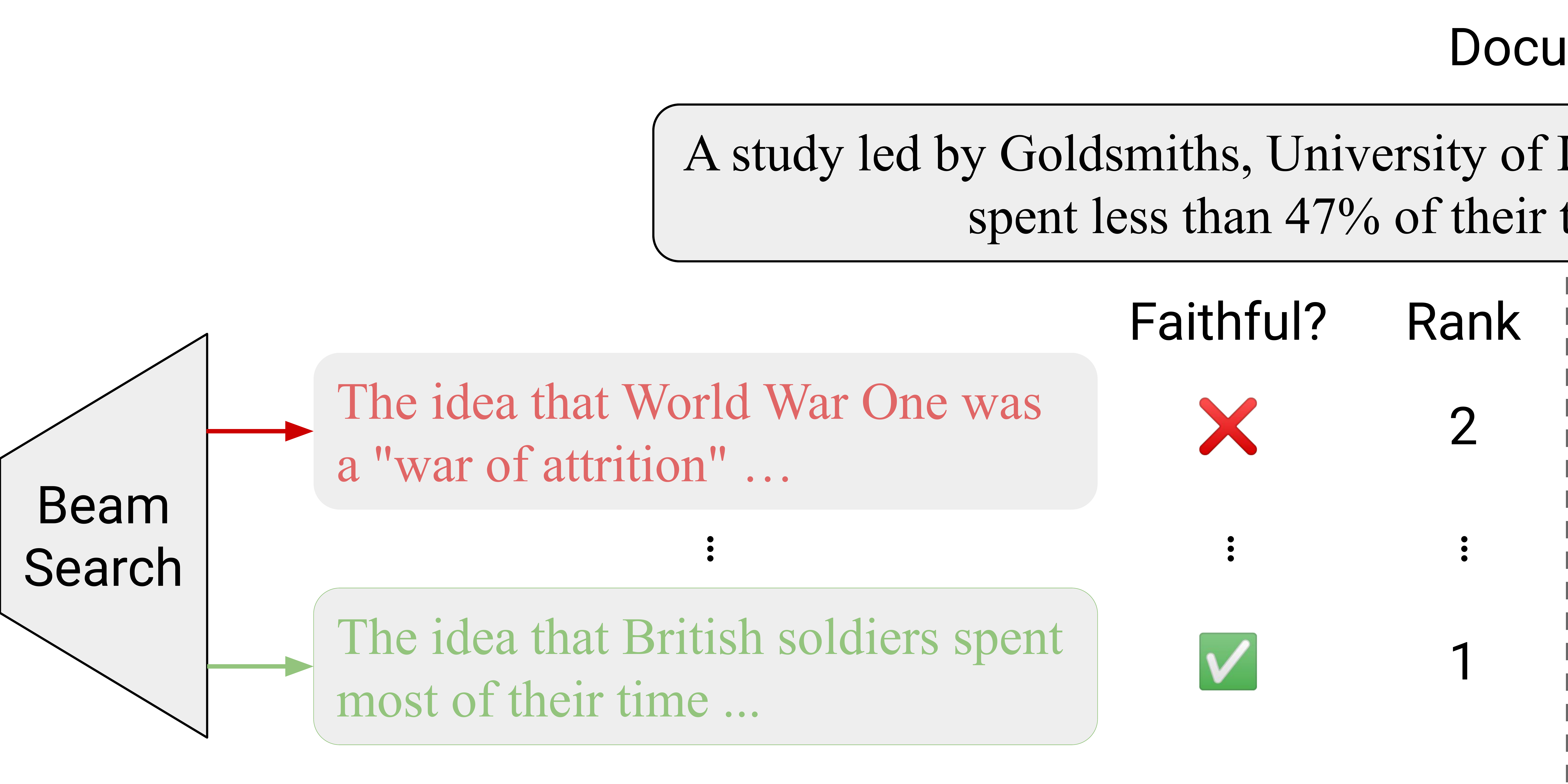}
    \caption{Ranker}
    \label{fig:ranking_figure}
    \end{subfigure}\begin{subfigure}{.5\textwidth}
    \centering
    \includegraphics[width=\linewidth]{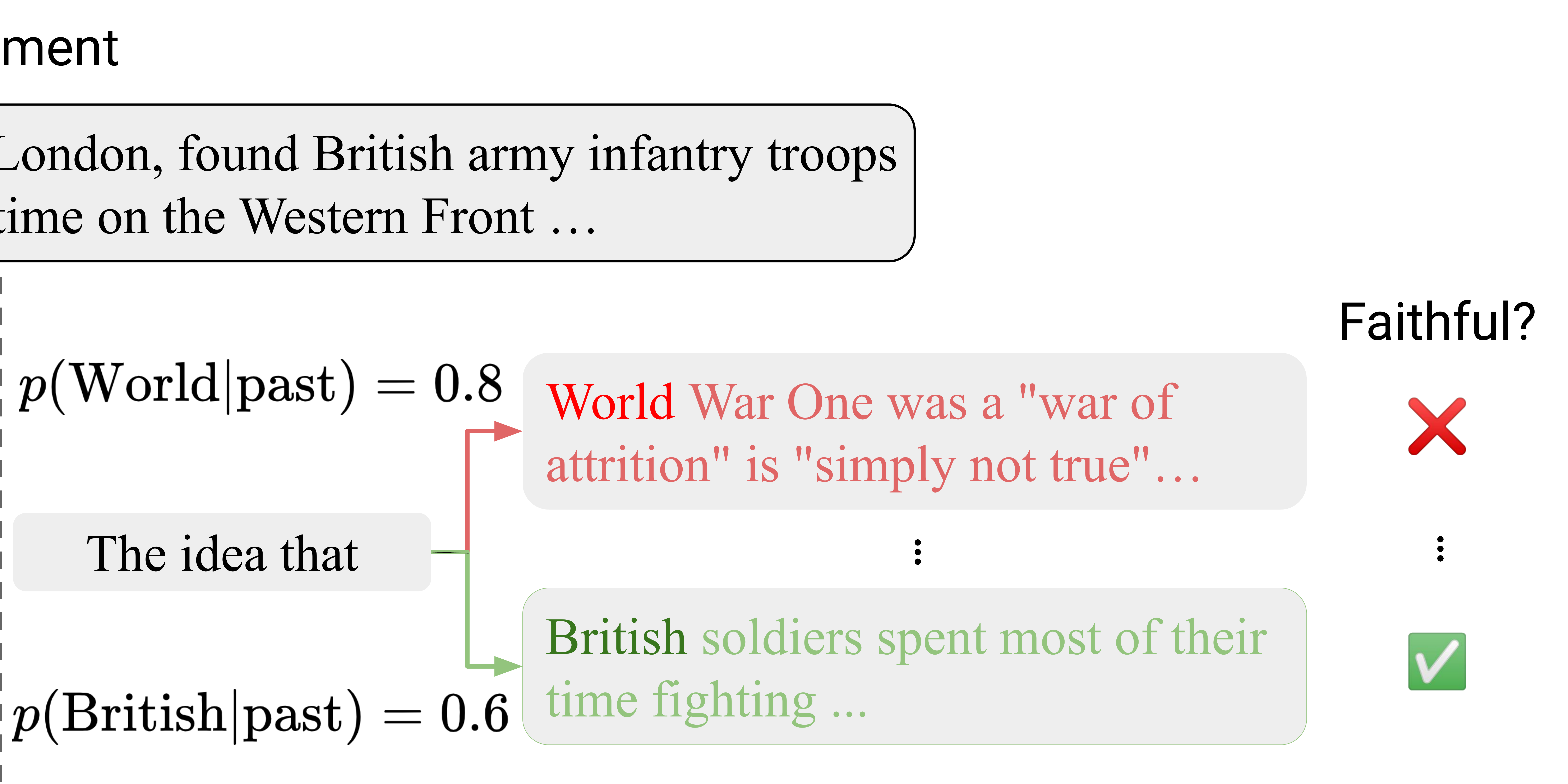}
    \caption{Lookahead}
    \label{fig:lookahead_figure}
    \end{subfigure}
    \caption{
    Illustration of our proposed decoding methods. \ref{fig:ranking_figure} shows our ranker that re-ranks the candidates produced by beam search according to faithfulness metrics. The first summary achieves a high score and would be used as the final summary for beam search, but it is not faithful. Our ranker ensures that the more faithful summary is ranked higher. \ref{fig:lookahead_figure} shows the lookahead heuristics that provide a faithfulness score given the full future summary. The model assigns a higher score to the word "World" than "British". However, by looking ahead we know that the completed summary following the most likely token will result in an unfaithful summary. Hence, the lookahead heuristics will ensure selecting the token "British" so that the resulting summary will be faithful.}
    \label{fig:method_figure}
\end{figure*}

\section{Introduction}
Recent developments in large pre-trained language models have achieved remarkable performance on abstractive summarization \cite{lewis-etal-2020-bart,zhang2020pegasus}. However, such models often suffer from the problem of hallucinations, where the generated summary contains facts or entities not present in the original document.
Prior research has analyzed and defined potential error types and typology \cite{maynez-etal-2020-faithfulness,pagnoni-etal-2021-understanding, poel-etal-2022-mutual}, and developed methods to improve faithfulness, including post-processing models \cite{chen-etal-2021-improving, dong-etal-2020-multi, liu-liu-2021-simcls,ladhak-etal-2022-faithful} and faithfulness-aware training \cite{goyal-durrett-2021-annotating, nan-etal-2021-entity, cao-wang-2021-cliff, wan-bansal-2022-factpegasus, zhang-etal-2022-improving-faithfulness,xiao-carenini-2022-entity}.

One aspect that is less understood on faithfulness of abstractive summarization is the effect of decoding strategies, which determine how the model generates the output strings.
Our primary objective is to understand whether different types of exploration of the search space, such as traversing and maintaining multiple possible output hypotheses with beam search or encouraging diversity 
with nucleus sampling \cite{holtzmann-etal-2020-curious}, have an impact on faithfulness.
To this end, we first conduct a thorough analysis comparing the faithfulness of popular decoding strategies, including greedy decoding, beam search, and nucleus sampling for two popular summarization datasets XSum \cite{narayan-etal-2018-dont} and CNN/DM \cite{hermann-etal-2015-teaching}. Evaluating the generated summaries using four faithfulness metrics, including BertScore \cite{zhang-etal-2020-bertscore}, FactCC \cite{kryscinski-etal-2020-evaluating}, DAE \cite{goyal-durrett-2021-annotating}, and QuestEval \cite{scialom-etal-2021-questeval}, and human evaluation, we find a consistent trend that beam search provides the most faithful summaries with its large exploration of the search space, and the randomness introduced by sampling hurts faithfulness.

To further improve faithfulness beyond the common decoding strategies, we propose two faithfulness-aware decoding methods. First, similar to \citet{falke-etal-2019-ranking}, we make use of the multiple candidates generated by beam search and propose a simple re-ranker, which selects the best summary according to a faithfulness metric. Instead of using a specific metric, we rank and select the summaries with a composite metric, a weighted combination of popular faithfulness metrics.
Next, inspired by \citet{lu-etal-2022-neurologic}, we propose a faithfulness heuristic that looks into the future to generate a full summary starting with the current tokens of any partially generated summary so as to provide a faithfulness score of the future summary during generation. The added heuristic ensures that the selected tokens will lead to a more faithful path in the search space. Compared to the baseline decoding strategies we analyzed, the two proposed methods significantly improve faithfulness as evaluated by four automatic faithfulness metrics and further confirmed by human evaluation.

Finally, to overcome the computational and runtime overhead of our proposed decoding methods, we explore distillation to transfer the knowledge of generating faithful summaries from a teacher model to a student model. Specifically, we use the faithfulness-aware decoding strategies as the teacher model to generate reference summaries. Then, we train student models, which have not been fine-tuned on the original task, to imitate the more faithful generation techniques
using an additional cross-entropy loss between the generated summaries by the student and teacher models.
Results indicate that the student model is able to generate summaries of similar faithfulness to that of the full teacher model while reducing the decoding time (seconds per example) up to $1/6$ of what the teacher model takes.
This process can be performed iteratively by using the student model as the teacher for the next iteration (See \autoref{fig:iterative_distillation}). With each iteration, the new student model is able to generate more faithful summaries, and outperform the original teacher model with just two iterations.

To summarize, our contributions are:
\begin{enumerate}
    \setlength{\itemsep}{-1pt}
    \item An analysis of the effect of popular decoding strategies, including greedy, beam, and nucleus sampling, on the faithfulness of abstractive summarization.
    \item Two faithfulness-aware generation methods, ranking and lookahead, that improve faithfulness over existing decoding strategies.
    \item A simple distillation approach that allows a student model to generate faithful summaries with just greedy decoding.
\end{enumerate}

\section{Faithfulness Behavior of Popular Decoding Strategies}\label{sec:faithfulness_analysis}
We first describe our experiment investigating the effect of popular decoding strategies on faithfulness. We wish to primarily investigate whether better exploration of the search space, such as the candidate expansion with beam search, can improve faithfulness, and how randomness introduced through sampling impacts faithfulness.
These investigations in turn motivate our more advanced, faithfulness-aware decoding strategies in Section~\ref{sec:faithful_decoding}.

\paragraph{Decoding Strategies (Greedy, Beam, and Nucleus Sampling).} For generation, we assume the common left-to-right, auto-regressive setting where the model generates a summary $y$ with $n$ tokens given the input document $x$:
\begin{equation*}
P(y|x) = \prod_{t=1}^{n}{p(y_t|y_{1:t-1},x)}
\end{equation*}
The summary tokens are selected with probability according to the decoding strategies.
We explore three common decoding strategies: greedy, beam search, and nucleus sampling \cite{holtzmann-etal-2020-curious}.
\textbf{Greedy search} selects the next token by the most probable token $y_{t} = \operatorname*{arg\,max}_{y} p(y|y_{1:t-1},x)$. \textbf{Beam search} extends greedy search by keeping top-$k$ hypothesis at each time step, where $k$ is the number of beams.
Another approach to decoding is to use sampling, where we consider \textbf{nucleus sampling}. \citet{holtzmann-etal-2020-curious} surprisingly find that methods that optimize probability, such as beam search, may lead to text degeneration, and thus propose nucleus sampling, a method that randomly selects from top tokens whose cumulative probability satisfies the threshold $p$. A small $p$ means less randomness and becomes greedy search, while a large $p$ allows for a more diverse output.

\section{Faithfulness-Aware Decoding Strategies}\label{sec:faithful_decoding}
We hypothesize (and later test and confirm whether it is true in Section~\ref{sec:baseline_results} and Appendix~\ref{sec:ablations_appendix}) that current decoding methods, such as beam search which explores a large space, may not explore the paths that focus on faithfulness directly and effectively.
Hence, we propose two faithfulness-aware methods that can be applied on top of the base decoding strategies to modify how the space is explored from two different perspectives:
(1) Ranking makes use of the large exploration of beam search and picks the explored path that is most faithful; (2) Lookahead directly guides the search process by adding faithfulness heuristics when selecting the next token starting from the initial decoding process.

\subsection{Ranking with Faithfulness Metrics}\label{sec:ranking}

Since beam search already explores many different suitable candidates during the decoding process, we hypothesize that more faithful summaries exist in the list of possible candidates, even if the model score is not directly optimized towards faithfulness (we show that this is true later in Section~\ref{sec:baseline_results}). Thus, we propose to rerank the generated candidates from beam search according to faithfulness metrics.

The process is illustrated in \autoref{fig:ranking_figure}. Assuming a beam search with beam size $k$, we have $k$ summaries generated by the decoding method. We compute a faithfulness metric (details of the metrics are presented in Section~\ref{sec:metrics}) over all summaries
and select the summary that achieves the highest faithfulness score. In the example, the more faithful summary that was originally ranked low according to model score is now ranked as the top summary according to faithfulness.

Re-ranking candidates for abstractive summarization have been studied primarily from the informativeness perspective \cite{ravaut-etal-2022-summareranker, ravaut-etal-2022-towards}, and our focus is on improving faithfulness. Our idea is most similar to \citet{falke-etal-2019-ranking}, where the authors use NLI models to re-rank. However, the results indicate that the NLI performance does not translate to improvement in faithfulness; their best-ranking model actually increases the number of unfaithful summaries at the top summary after re-ranking by 3\%. The authors attribute it to domain shift and NLI models relying on simple heuristics like lexical matching. We thus explore using faithfulness metrics directly for ranking.

\paragraph{Composite Metric.} While it is possible to use one of the faithfulness metrics to rank the candidates, it often leads to over-fitting for one particular metric (each metric can have its own domain biases and idiosyncrasies) and hurts the overall faithfulness scores evaluated by other metrics. We instead tune a composite metric that aggregates the vote of several popular metrics (See Section~\ref{sec:decoding_details}).
We use linear regression to provide weights for each metric and tune on human judgments of faithfulness.
We refer the readers to Appendix~\ref{sec:composite_metric_appendix} and Appendix~\ref{sec:ablations_appendix} for details and ablations for the composite metric.

\subsection{Lookahead}
\citet{lu-etal-2022-neurologic} use lookahead to provide a future constraint satisfaction estimate and show its effectiveness in several constrained generation tasks (commonsense generation, constrained machine translation, table-to-text generation, and constrained question generation).
We extend this idea to improve faithfulness of abstractive summarization. Instead of relying on explicit constraints that are available for the constrained generation tasks, we use reference-free faithfulness metrics on the full future summaries as an estimate. Unlike re-ranking which is constrained by the search space explored by beam search, lookahead allows for exploration of a much larger number of candidates.

Figure~\ref{fig:lookahead_figure} shows an example of the lookahead. When selecting the next token, the usual decoding scheme would select the word "World" that has the highest probability. However, if we were to follow this path, the resulting summary would introduce hallucinations. Instead, we would like to guide the model to select the less probable token "British," which will yield a faithful summary sentence.

Formally, each summary token is selected by:
\begin{equation*}
f(y_t) = \log{P(y_{1:t} \mid x)} + w \cdot \max_{L_{y \leq t}} {h(y_{1:t+l},x)}
\end{equation*}
where $\log{P(y_{1:t} \mid x)}$ is the model score, $h(\cdot)$ is a reference-free faithfulness evaluation function that assigns a score to the summary, $w$ is the weight, and $l$ is the number of tokens to look into the future.

Here, $L_{y \leq t }$ is a set of possible generated summaries that start with the summary tokens $y_{1:t}$. The number of summaries for $L$ varies given the decoding strategies we use to generate future summaries. Greedy search and sampling produce a single expansion, and beam search produces $k$ number of summaries depending on the beam size.
Although the lookahead length $l$ can be specified, we instead generate the full summary, as current faithfulness metrics expect full summaries as input and do not work well on partial summaries (see Appendix~\ref{sec:ablations_appendix}).

\subsection{Combining Ranking and Lookahead}
We can combine the two methods to further improve faithfulness. We first use the \lookahead{} to generate faithful beam candidates and then select the best candidates with ranking. We refer to this method as \textsc{Beam+Lookahead+Ranking}.

\begin{figure}[!t]
    \centering
    \includegraphics[width=.9\columnwidth]{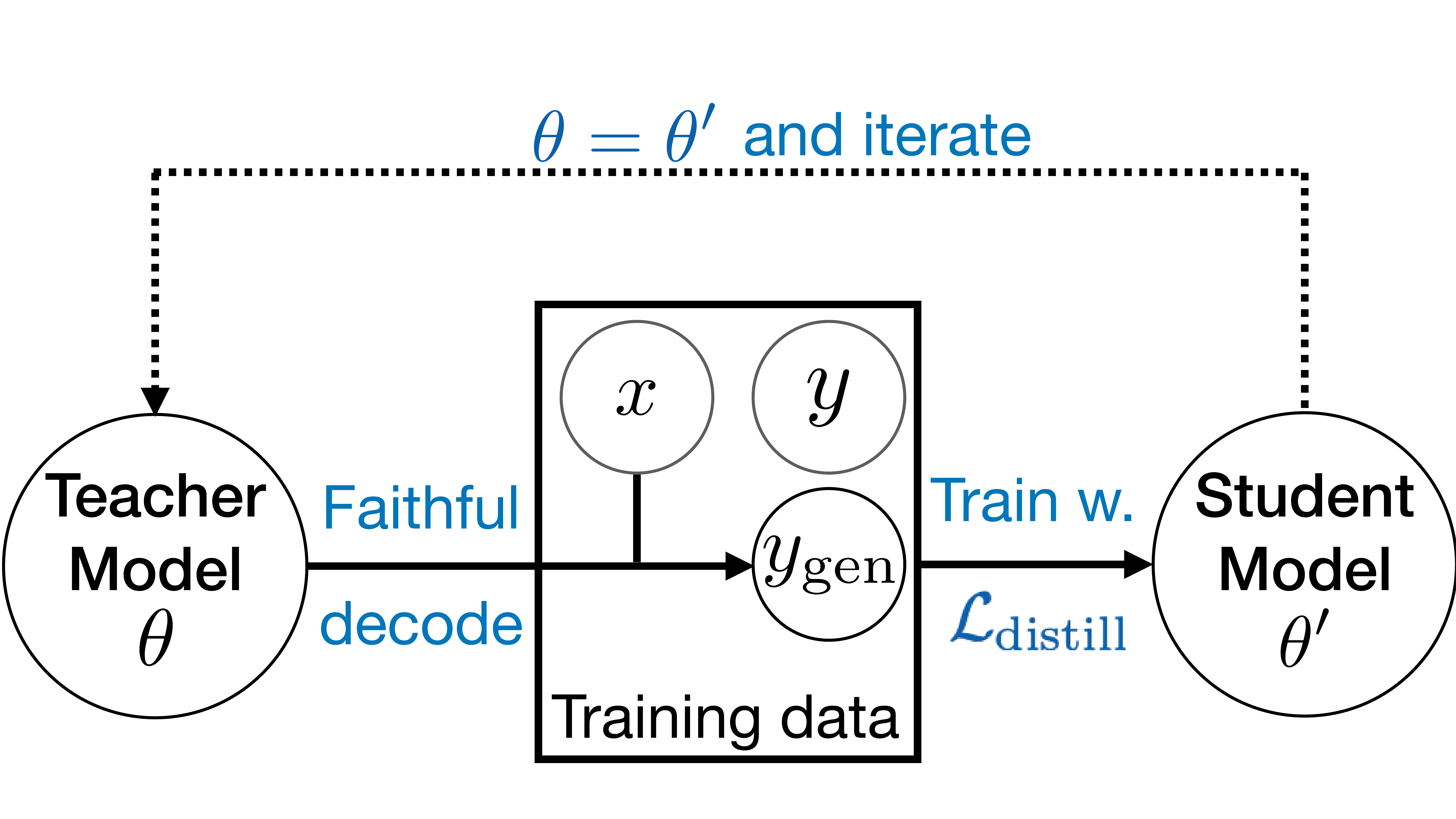}
    \caption{Illustration of the iterative distillation process.
    We train a student model $\theta'$ with summaries generated by the teacher model $\theta$, which uses faithfulness-aware decoding methods.
    The resultant student model $\theta'$ that is trained on more faithful summaries can in turn be used as $\theta$ to generate the training data for the next iteration.
    }
    \label{fig:iterative_distillation}
\end{figure}

\section{Efficient Decoding via Distillation} \label{sec:distillation}
One drawback of the proposed decoding methods is the heavy computational cost during decoding. We thus explore using distillation to transfer the knowledge of faithfulness-aware decoding to a student model that can generate summaries of similar faithfulness with just greedy decoding. We note here that our distillation aims at improving the decoding time rather than downsizing the model.
Similar to \citet{kim-rush-2016-sequence}, we assume that we have a teacher model and a student model. In our setting, the teacher model does not necessarily need to be a different model, but it needs to decode with more faithfulness-aware methods. Typical distillation methods use the teacher's probability distribution \cite{kim-rush-2016-sequence} as the target for the student model to imitate. In our case, however, that distribution is the same for all methods -- the difference lies in how the probability is used to generate the next tokens. Thus, we propose a new decoding distillation loss. We use the teacher model to generate summaries $y_{\text{gen}}$ as additional reference summaries, and interpolate between the cross-entropy loss using the original reference summaries and the cross-entropy loss where we consider $y_{\text{gen}}$ as reference summaries.
Formally, the training loss is:
\begin{equation*}
\mathcal{L}_{\text{distill}} = \mathcal{L}_{\text{XE}}(y',y) + \lambda \mathcal{L}_{\text{XE}}(y',y_{\text{gen}})
\end{equation*}
where $\mathcal{L}_{\text{XE}}$ is the cross entropy, $y'$ is the generated summary by the student model, and $\lambda$ is a hyper-parameter for the weight of the cross-entropy loss on the generated summaries.

\paragraph{Iterative Distillation.}
While we use the student model with just greedy decoding to improve decoding speed, the student model can also benefit from using our proposed faithfulness-aware decoding methods. Thus, the student models can also serve as a new teacher model to distill more faithfulness knowledge to a new student model. The distillation process thus becomes iterative, illustrated in \autoref{fig:iterative_distillation}. We use the trained student model as a new teacher model, where we decode with our proposed faithfulness methods to create additional reference summaries $y_{\text{gen}}$ for the next iteration.

\section{Experiments}

\subsection{Datasets and Models}\label{sec:dataset_models}
We perform experiments on two popular datasets for abstractive summarization, XSum \cite{narayan-etal-2018-dont} and CNN/DM \cite{hermann-etal-2015-teaching}. More details on the datasets are described in Appendix~\ref{sec:dataset_appendix}. We use the released checkpoint of BART-large (406M) for the two datasets.\footnote{We use the checkpoint \textsc{bart-large-xsum} (\url{https://huggingface.co/facebook/bart-large-xsum}) and \textsc{bart-large-cnn} (\url{https://huggingface.co/facebook/bart-large-cnn}).}
The same experiment is done with PEGASUS \cite{zhang2020pegasus}, which is presented in Appendix~\ref{sec:analysis_appendix}.

\subsection{Evaluation Metrics}\label{sec:metrics}
We use the F1 measure of ROUGE-L \cite[RL]{lin-2004-rouge}, i.e., the overlap of the longest common subsequence between a generated summary and reference summary, and the F1 measure of BERTScore \cite[BS]{zhang-etal-2020-bertscore} to evaluate summary quality. In addition, we use BS-Fact, i.e., the BERTScore precision of a summary with respect to its source document rather than the reference summary, FactCC \cite{kryscinski-etal-2020-evaluating}, DAE \cite{goyal-durrett-2021-annotating}, and QuestEval \cite{scialom-etal-2021-questeval} for faithfulness evaluation. Details of the metrics are presented in section~\ref{sec:experiment_detail_appendix}.

\begin{figure*}
    \centering
    \includegraphics[width=\textwidth]{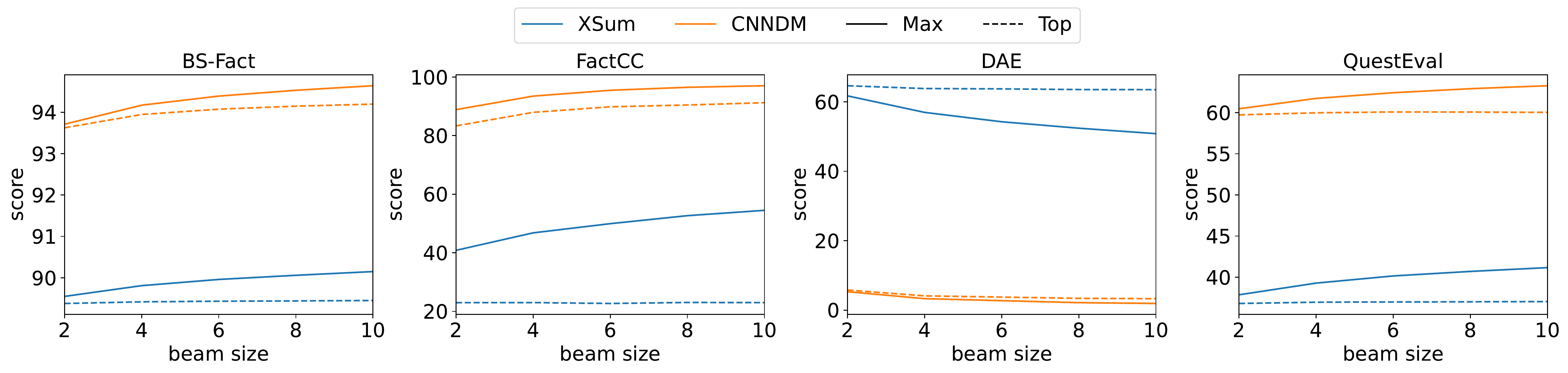}
    \caption{
    Maximum possible score (Max) for each faithfulness metric and the faithfulness scores of the top candidate (Top) at various beam sizes. As beam size increases, more faithful summaries exist in the list of candidates, but the faithfulness of the top beam improves only slightly.
    }
    \label{fig:beam_max}
\end{figure*}

\subsection{Human Evaluation Setup}

We use Amazon Mechanical Turk (AMT) to ask human annotators to judge the faithfulness and informativeness of the summaries generated with different decoding.

\paragraph{Faithfulness.}
We ask workers to judge the faithfulness of a summary sentence using a 3-star rating (1=major factual error, 2=minor factual error, 3=no factual error. Three judgments per summary are then aggregated using majority voting. We randomly select 200 examples from both datasets and use the summaries generated using greedy, sampling, beam search, as well as the ranking and lookahead strategies applied to beam search.  We report the percentage of summaries that are fully factual (i.e. the percentage of summaries rated as 3-star) as the faithfulness score, and also report the distribution of summaries rated as 1, 2, and 3 stars. Details on qualification, payment and other aspects of the evaluation can be found in Appendix~\ref{sec:human_evaluation_appendix}.

\paragraph{Informativeness.}
We also evaluate the generated summaries in terms of informativeness. We consider summary to be informative if its content is important and relevant, but it does not necessarily need to be long.
We use best-worst-scaling (BWS) for evaluating the informativeness of the generated summaries, as this method is “a less labor-intensive alternative to paired comparisons that has been shown to produce more reliable results than rating scales” \cite{kiritchenko-mohammad-2017-best}.
Accordingly, for each dataset, we select 200 random articles with the corresponding summaries from five systems in random order. We ask three annotators to select the most informative (“best”) and the least informative (“worst”) among the five. A rating per system is computed as the percentage of times it is chosen as best minus the percentage of times it is selected as worst. A value of 100 means that the system has been unanimously picked as “best”, whereas a value of -100 means that the system has been unanimously picked as “worst”. Additional details, as well as  
the screenshot of the annotation interface, are in Appendix~\ref{sec:human_evaluation_appendix}.

\subsection{Decoding Setting Details}\label{sec:decoding_details}
We describe the settings of the basic decoding methods, our faithfulness-aware decoding methods and distillation. More details are in Appendix~\ref{sec:decoding_details_appendix}.

\paragraph{Basic Decoding Method.} We compare the summaries generated using greedy search, beam search ($k=10$), and nucleus sampling ($p=0.9$). Additional experiments with various beam sizes and top-p values can be found in Appendix~\ref{sec:analysis_appendix}.

\paragraph{Ranking and Composite Metric.} We use beam search ($k=10$) and rank the candidates using the composite metric introduced in Section~\ref{sec:ranking}. To train the composite metric, we explore combining FactCC, BS-Fact, DAE, and QuestEval. We use \textsc{FactCollect} \cite{ribeiro-etal-2022-factgraph}, a large collection of four faithfulness annotations to train a linear regression on the human-labeled faithfulness judgments. More details of the composite metric and its robustness to another domain can be seen in Appendix~\ref{sec:composite_metric_appendix}.

\paragraph{Lookahead.} We use BS-Fact as the faithfulness metric for the lookahead as it correlates highly with human judgment \cite{pagnoni-etal-2021-understanding} and is quick to compute without the need for additional pre-processing. We use greedy search to generate future summaries and apply it to both greedy and beam searches.

\paragraph{Distillation.} We use the checkpoint of our two proposed faithfulness-aware decoding methods as the teacher model, and train the student model from \textsc{BART-large}.\footnote{\url{https://huggingface.co/facebook/bart-large}} We follow the original fine-tuning hyperparameters provided by the authors \cite{lewis-etal-2020-bart} and use $\lambda=1$ for the weight of the additional cross-entropy loss.

\begin{table}[!t]
    \centering
    \resizebox{0.99\columnwidth}{!}{\begin{tabular}{l c c c c c c}
    \toprule
& RL & BS & BS-Fact & FactCC & DAE $\downarrow$ & QuestEval \\
    \midrule
    \multicolumn{7}{c}{CNN/DM} \\
    \midrule
    Greedy & \textbf{30.93} & \textbf{88.39} & 93.15 & 69.61 & 8.15 & 59.13 \\
    Nucleus & 27.64 & 87.90 & 91.76 & 54.05 &  21.61 & 56.43 \\
    Beam & 29.99 & 88.03 & \textbf{94.20}  & \textbf{84.23} & \textbf{3.30} & \textbf{60.03} \\
    \midrule
    \multicolumn{7}{c}{XSum} \\
    \midrule
    Greedy &  36.16 & 92.03 & 89.28 & \textbf{23.53} &  65.35 & 36.51 \\
    Nucleus & 31.15 & 91.26 & 88.62 & 21.04 & 76.20 & 34.98 \\
    Beam & \textbf{37.11} & \textbf{92.12} & \textbf{89.45} & 22.97 & \textbf{63.49} & \textbf{37.05}\\
    \bottomrule
    \end{tabular}
    }
    \caption{Baseline results of popular decoding methods
    measured by summarization quality metrics
    (Rouge-L (RL) and BertScore (BS)) and faithfulness metrics. We observe a general trend where beam search performs the best and nucleus sampling performs the worst in terms of faithfulness.
    Full result with different beam sizes and top-$p$ probability for nucleus sampling is in \autoref{tab:analysis_full}.
    }
    \label{tab:analysis}
\end{table}

\begin{table*}[!t]
    \centering
    \small
    \begin{tabular}{l c c c c c c}
    \toprule
    & Rouge-L & BERTScore & BS-Fact & FactCC & DAE $\downarrow$  & QuestEval \\
    \midrule
    \multicolumn{7}{c}{CNN/DM} \\
    \midrule
    Greedy &  \textbf{30.93} & \textbf{88.39} & 93.15 & 69.61 & 8.15 & 59.13 \\
    Beam &  29.99 & 88.03 & 94.20 & 84.23 & 3.30 & 60.03 \\
    \ranking{} & 30.08 & 88.12 & 94.31 & 90.27 & 1.92 & 62.57 \\
    \textsc{Greedy+Lookahead} & 30.75 &	88.35 & 93.90 & 71.54 & 5.70 & 60.13 \\
    \lookahead{} & 28.66 & 87.84 & \textbf{95.32} & 86.10 & 1.68 & 61.80 \\
    \textsc{Beam+Lookahead+Ranking} & 28.86 & 87.92 & 95.26 & \textbf{91.68} & \textbf{1.08} & \textbf{63.69} \\
    \midrule
    \multicolumn{7}{c}{XSum} \\
    \midrule
    Greedy & 36.16 & 92.03 & 89.28 & 23.53 & 65.35 & 36.51 \\
    Beam &  \textbf{37.11} & \textbf{92.12} & 89.45 & 22.97 & 63.49 & 37.05 \\
    \ranking{} & 36.42 & 92.10 & 89.79 & \textbf{40.11} & 51.48 & 40.10 \\
    \textsc{Greedy+Lookahead}  & 36.25 & 92.11 & 89.71 & 24.21 & 60.46 & 37.17 \\
    \lookahead{} & 35.27 & 91.94 & \textbf{90.78} & 23.38 & 50.04 & 39.24 \\
    \textsc{Beam+Lookahead+Ranking} & 34.71 & 91.90 & \textbf{90.78} & 38.86 & \textbf{41.04} & \textbf{41.94} \\
    \bottomrule
    \end{tabular}
    \caption{Results for our proposed decoding strategies. Compared to the baseline methods (greedy and beam search), both ranking and lookahead improve faithfulness. The combination of both methods further increases faithfulness.
    }
    \label{tab:full_result}
\end{table*}

\section{Results}

\subsection{Baseline Decoding Results}\label{sec:baseline_results}
We show the analysis of common decoding strategies in \autoref{tab:analysis}.
Both datasets show a similar trend. Beam search performs the best in terms of faithfulness except for FactCC on the XSum dataset. Compared to greedy decoding, which is beam search with $k=1$, the candidate expansion with a larger beam size provides better exploration for faithfulness.
Nucleus sampling degrades faithfulness compared with greedy search, showing that the introduced randomness is not helpful for faithfulness. This aligns with observations from \citet{narayan-etal-2022-well} and \citet{chen-etal-2021-wikitablet}, which show that nucleus sampling produces less relevant text for data-to-text generation.

The results are surprisingly mixed for both datasets in terms of summary quality, i.e., RL and BS scores. Comparing beam search with greedy decoding, we see improvement of both scores on XSum but not for CNN/DM. Nucleus sampling, on the other hand, is also worse than greedy search on this aspect, suggesting that randomness may not be suited for the task of abstractive summarization.

\paragraph{Search Space for Beam Search.} Inspired by  \citet{xu-etal-2022-massive} who hinted at the potential of better faithfulness with a large exploration of the search space, we use beam search to explore whether larger beam sizes (and hence larger exploration) derive more faithful summaries. To this end, we use all summaries generated by beam search and select the beam that would result in the highest possible score for each metric. We show the maximum score (Max) for the four faithfulness metrics and the faithfulness score of selecting the top beam (Top) given different beam sizes in \autoref{fig:beam_max}. We see a clear trend that increasing the beam size improves all faithfulness scores.
 This confirms our hypothesis that larger exploration of the search space can provide additional faithfulness gain, and thus showing the potential of our proposed decoding strategies, especially our reranking strategy, to output more faithful summaries. The faithfulness scores of TOP only increase marginally compared to the increase for Max, showing the importance of having better faithfulness guidance, such as our proposed faithfulness lookahead heuristics. 

\subsection{Faithfulness-Aware Decoding Results}
We now show the impact of faithfulness-aware methods compared with the traditional decoding methods, which is shown in \autoref{tab:full_result}. We first observe that applying ranking on top of beam search improves faithfulness significantly over beam search, as measured by all faithfulness metrics. Specifically, QuestEval reaches 62.57 (2.5 points improvement) and 40.10 (3.1 points improvement) on CNN/DM and XSum respectively. DAE error rate reduces from 63.49 to 51.48 and 3.30 to 1.92, which is a relative improvement of 18.92\% (12.01 points) and 41.8\% percent (1.38 points) on XSum and CNN/DM, respectively.

We observe similar improvement for lookahead as well, where applying the lookahead improves the faithfulness over the base decoding strategy over all faithfulness metrics. Nevertheless, the base decoding strategy is still the dominating factor, as \lookahead{} generates more faithful summaries than \textsc{greedy+lookahead}
for all faithfulness metrics. \textsc{Greedy+Lookahead} outperforms Beam on the XSum dataset, showing that better guidance with future faithfulness heuristics can improve faithfulness without large exploration.
Finally, the combination of lookahead and ranking can further improve faithfulness as evaluated by FactCC, DAE, and QuestEval.

In terms of ROUGE score, applying faithful decoding methods decreases RL. This tradeoff between faithfulness and ROUGE has been observed in many prior works \cite{chen-etal-2021-improving, kryscinski-etal-2020-evaluating,wan-bansal-2022-factpegasus}. One reason for this phenomenon is that more than 70\% of the reference summaries contain hallucinations \cite{maynez-etal-2020-faithfulness}, so the more faithful summaries that do not contain such hallucinations will have lower ROUGE scores. To investigate this problem, we perform a human evaluation study, where we find that the summaries generated by \lookahead{} are considered to be most informative. More details are in Appendix~\ref{sec:human_evaluation_appendix}.

\subsection{Human Evaluation Results}

\paragraph{Faithfulness.} The observation on automatic faithfulness metrics aligns with the result of human evaluation in \autoref{tab:human_eval_result}. For XSum, among the baseline decoding methods, we see that sampling performs the worst. Interestingly, greedy is more faithful than beam search, but the difference is only 1.5 points. Our proposed decoding strategies generate summaries that are judged more faithful compared to that of the baseline decoding strategies. Specifically, \lookahead{} reaches 56.5, even outperforming \ranking{} by 5 points.
We also observe that our proposed methods are able to significantly reduce the percentage of summaries that are considered to contain major factual errors; Compared to beam search, ranking reduces the percentage from 44.5 to 36.5,
and lookahead further reduces the percentage by 3 points. For CNN/DM, we see the striking result that the summaries generated by our proposed methods achieve the highest faithfulness, and among the two systems, there are no major errors for \lookahead{}.

\paragraph{Informativeness.}
The result is shown in \autoref{tab:human_eval_informativeness}. The output of the \lookahead{} is clearly seen as the most informative among the five methods. This result suggests that Rouge-L and BERTScore may not be good indicators for informativeness, as \lookahead{} achieves the lowest scores for the two automatic metrics on both datasets.

\begin{table}[!t]
    \centering
    \resizebox{0.99\columnwidth}{!}{\begin{tabular}{c c c c | c c c}
     \toprule
     & \multicolumn{3}{c|}{XSum} & \multicolumn{3}{c}{CNN/DM} \\
      & 1 & 2 & 3 & 1 & 2 & 3 \\
      \midrule
      Greedy &  43.0 & 12.0 & 45.0 & 4.0 & 3.5 & 92.5 \\
      Sampling & 55.0 & 13.0 & 32.0 & 8.0 & 10.5 & 81.5 \\
      Beam & 44.5 & 12.0 & 43.5 & 0.0 & 2.0 & 98.0 \\
      \ranking{} & 36.5 & 14.0 & 49.5 & 0.5 & 1.0 & \textbf{98.5} \\
      \lookahead{} & 31.5 & 12.0 & \textbf{56.5} & 0.0 & 1.5 & \textbf{98.5}  \\
    \bottomrule
    \end{tabular}
    }
    \caption{Human evaluation results on faithfulness with the 3-star rating system (1=major factual error, 2=minor factual error, 3=no factual error). Our proposed faithfulness-aware methods are judged as the most faithful (the percentage of summaries rated as 3), confirming our observation with automatic faithfulness metrics.
    }
    \label{tab:human_eval_result}
\end{table}

\subsection{Abstractiveness}\label{sec:abstractiveness}

\begin{figure}[!t]
    \centering
    \includegraphics[width=.95\columnwidth]{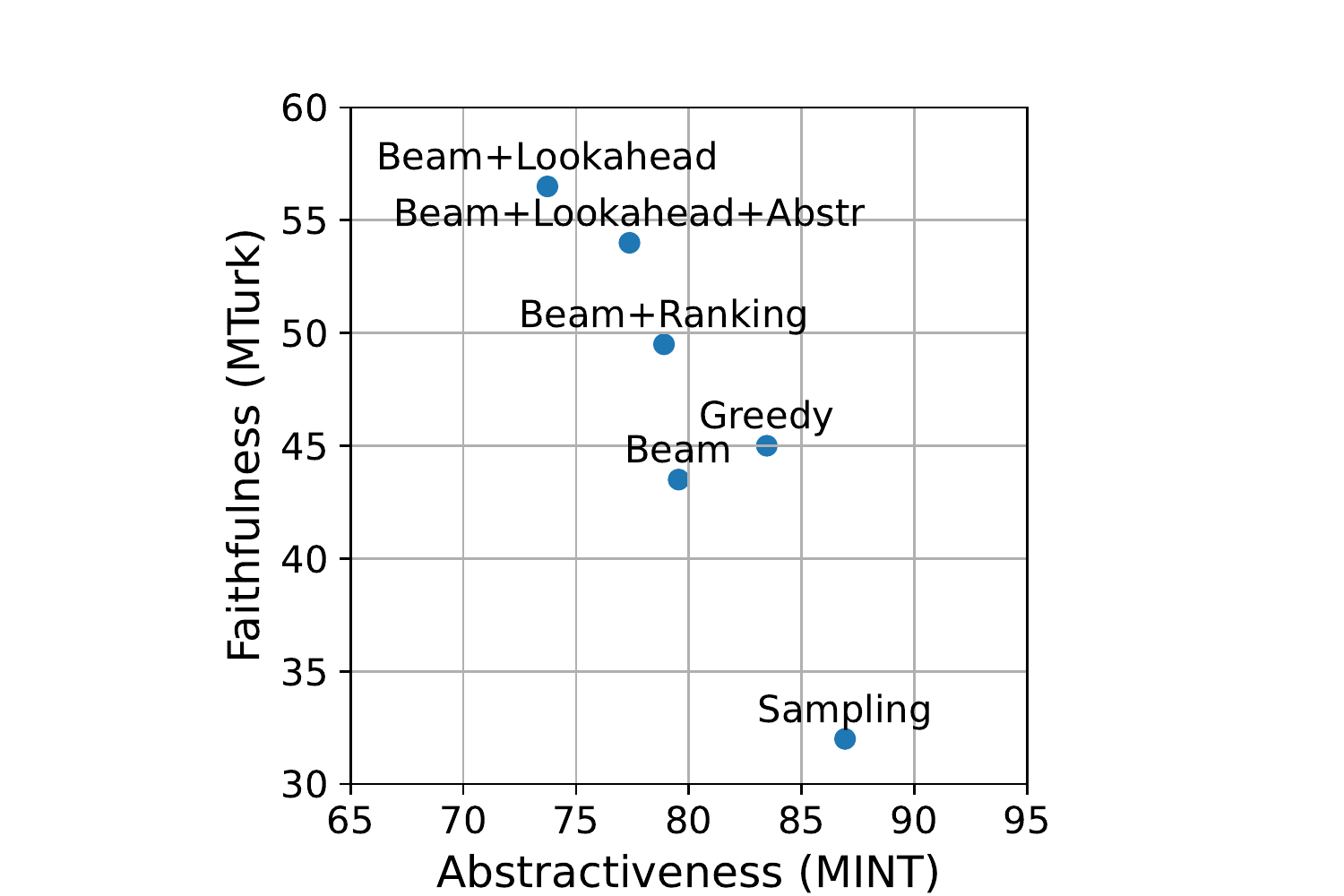}
    \caption{Faithfulness and abstractiveness tradeoff results on 200 XSum examples used for human annotation.
    \textsc{Beam+Lookahead+Abstr} is the model that is trained with additional abstractiveness heuristics (See Section~\ref{sec:abstractiveness} for more details).
    }
    \label{fig:abstractiveness}
\end{figure}

Models can "trivially" become more faithful by becoming more extractive \cite{dreyer-etal-2020-analyzing}, and thus it is important to understand where the gain in faithfulness stems from. We experiment on XSum, as methods can achieve larger improvement in faithfulness and thus potentially more gain through extensiveness. We experiment with the 200 examples used for human evaluation and calculate \textsc{Mint} \cite{dreyer-etal-2020-analyzing} 
for abstractiveness and plot this score against the human-labeled faithfulness, similar to \citet{ladhak-etal-2022-faithful}.
The result is shown in \autoref{fig:abstractiveness}.
Similar to the observation of \citet{dreyer-etal-2020-analyzing}, more faithful models tend to be more extractive; however, the gain in faithfulness is considerably larger than the decrease in abstractiveness. For example, comparing \lookahead{} with beam search, the relative increase in faithfulness ($29.89\%$) is quadruple the decrease ($7.27\%$) in abstractiveness.
Similar experiments on CNN/DM are in Appendix~\ref{sec:abstractiveness_appendix}.

\paragraph{Lookahead with Faithfulness and Abstractiveness.} We further show that our lookahead method can easily allow additional heuristics, such as balancing both faithfulness and abstractiveness. Specifically, we replace $h(\cdot)$ with combination of BS-Fact and \textsc{Mint}:
\begin{equation*}
h(y,x) = \alpha \text{BS-Fact}(y,x) + (1-\alpha) \textsc{mint}(y,x)
\end{equation*}
We use $\alpha=0.75$ and the same hyper-parameters as \lookahead{}. We refer to this model as \textsc{Beam+Lookahead+Abstr} and show the point in \autoref{fig:abstractiveness}. Compared to \lookahead{}, this model can increase abstractiveness at a small cost in faithfulness, demonstrating the flexibility of our lookahead method to incorporate various characteristics for summarization.

\begin{table}[!t]
    \centering
    \small
    \begin{tabular}{c c|c}
    \toprule
     & XSum & CNN/DM \\
     \midrule
     Greedy & $\phantom{-0}3.0$ & $-\phantom{0}8.2$ \\
     Sampling & $-20.5$ & $-23.8$ \\
     Beam & $\phantom{-0}1.8$ & $\phantom{-0}8.5$ \\
     \ranking{} & $\phantom{-0}1.0$ & $-\phantom{0}2.8$ \\
     \lookahead{} & $\phantom{-}\textbf{17.7}$ & \textbf{$\phantom{-}\textbf{31.0}$} \\
     \bottomrule
    \end{tabular}
    \caption{Human evaluation results on informativeness with best-worst-scaling ($100$=unanimous best, $-100$=unanimous worst).}
    \label{tab:human_eval_informativeness}
\end{table}

\subsection{Distillation}

We present the distillation result in \autoref{tab:distillation_result}. While the student models are not able to outperform the teacher models, they approach the performance of the teacher models. The student models are also able to generate more faithful summaries compared to the greedy search baseline, which is only trained using the cross-entropy loss $\mathcal{L}_{\text{XE}} (y',y)$.

The main benefit of the student model comes from the improved decoding speed. The ranking time reduces from 0.77 seconds per example to 0.47, which is a $40\%$ improvement. The largest gain can be seen for lookahead, where the decoding speed reduces from 3 seconds per example to 0.49, only $1/6$ of the time it was originally taking.

For example, the student model distilled from \ranking{} improves DAE by 6.6 points and QuestEval by a point compared to the greedy search baseline and only differs from the teacher model by 2.5 points for DAE and 0.5 points for QuestEval. When using a more faithful teacher model, i.e. \lookahead{}, the student model is able to generate more faithful summaries, as evaluated by BS-Fact, DAE, and QuestEval.

\paragraph{Iterative Distillation.} Next, we show the result of distilling \ranking{} iteratively on XSum in \autoref{tab:iterative_result}.  We see that with each iteration, the model is able to improve faithfulness further. When compared to the original teacher model, \ranking{}, the student model is able to outperform all faithfulness metrics with two iterations. We stress that here all models only use greedy decoding, thus showing the potential of combining decoding with training for more faithful models.

\begin{table}[!t]
    \centering
    \resizebox{\columnwidth}{!}{\begin{tabular}{l c c c c c c c}
    \toprule
    & RL & BS & BS-Fact & FC & DAE $\downarrow$  & QE & Speed \\
    \midrule
    Greedy & 36.16 & 92.03 & 89.28 & 23.53 & 65.35 & 36.51 & 0.39 \\
    \midrule
    \multicolumn{8}{c}{\ranking{}}\\
    \midrule
    Teacher  & 36.46 & \textbf{92.14} & \textbf{90.15} & \textbf{22.47 }& \textbf{56.33} & \textbf{37.98} & 0.77 \\
    Student & \textbf{36.59} & 92.07 & 89.80 & 23.52 & 58.78 & 37.46 & \textbf{0.47} \\
    \midrule
    \multicolumn{8}{c}{\lookahead{}} \\
    \midrule
    Teacher & 35.91 & 92.06 & \textbf{90.51} & 22.44 & \textbf{52.50} & \textbf{38.72} & 3.00 \\
    Student & \textbf{36.52} & \textbf{92.07} & 89.97 & \textbf{22.58} & 58.02 & 37.89 & \textbf{0.49} \\
    \bottomrule
    \end{tabular}
    }
    \caption{
    Distillation results using our proposed faithfulness-aware decoding methods as the teacher. We abbreviate FactCC as FC and QuestEval as QE. Speed is calculated by seconds per summary.
    }
    \label{tab:distillation_result}
\end{table}

\section{Related Work}

Many of the related works of our proposed decoding methods have been discussed in Section~\ref{sec:faithful_decoding}; here we cover other related areas.

\paragraph{Decoding methods.} A decoding  method for text generation explores an approximate search method to select the best tokens to form a hypothesis. Several works have critically analyzed different decoding strategies for  natural language generation, including  beam search \cite{meister-etal-2020-beam,stahlberg-byrne-2019-nmt,xu-etal-2022-massive,holtzmann-etal-2020-curious}, best-first-search \cite{meister-etal-2020-best}, and lattice \cite{xu-etal-2022-massive}. While these works investigated the effectiveness of decoding methods on generated outputs from the perspective of diversity and repetitiveness, to our best knowledge, none of the works have explicitly analyzed their performance on faithfulness.

\paragraph{Distillation.} Distillation aims at compressing the knowledge from a larger model into a smaller one. A conventional approach uses soft targets, i.e. learning the logits of a teacher model rather than final predictions \cite{bucila-etal-2006-model,hinton-etal-2015-distilling,kim-rush-2016-sequence}. While this method has shown to be very effective, it is less applicable to our case where the underlying distribution for the next probable tokens does not necessarily change (for ranking, we do not modify the model scores at all)
and thus not useful to learn soft labels. Different from compressing model size, our approach focuses on reducing the computational cost during decoding.
Our method is most similar to pseudo-labeling \cite{shleifer-rush-2020-pre}, where we use generated summaries as "hard" labels. We do not replace reference summaries with our generated ones. Instead, we use interpolation \cite{kim-rush-2016-sequence} to account for both faithfulness and quality.

\begin{table}[!t]
    \centering
    \resizebox{\columnwidth}{!}{\begin{tabular}{l c c c c c c}
    \toprule
    & RL & BS & BS-Fact & FactCC & DAE $\downarrow$  & QuestEval \\
    \midrule
    Teacher  & 36.46 & \textbf{92.14} & 90.15 & 22.47 & 56.33 & 37.98 \\
    Iter. 1 & \textbf{36.59} & 92.07 & 89.80 & 23.52 & 58.78 & 37.46 \\
    Iter. 2 & 35.95 & 91.95 & 90.16 & 23.14 & 54.01 & 38.10 \\
    Iter. 3 & 35.09 & 91.73 & 90.48 & 22.77 & 50.66 & 38.86 \\
    Iter. 4 & 34.32 & 91.54 & 90.81 & 24.49 & 47.83 & 39.64 \\
    Iter. 5 & 33.60 & 91.34 & \textbf{91.11} & \textbf{25.52} & \textbf{45.85} & \textbf{40.39} \\
    \bottomrule
    \end{tabular}
    }
    \caption{Iterative distillation results using \ranking{} as the teacher decoding  method. With two iterations, the  student model is able to outperform the original teacher model in  terms of faithfulness, and further iterations continuously improve faithfulness.}
    \label{tab:iterative_result}
\end{table}

\section{Conclusion}
In this paper, we show a thorough analysis of the effect of decoding strategies on faithfulness for abstractive summarization. We present an analysis of popular decoding strategies, as well as our two newly proposed faithfulness-aware decoding strategies, ranking and lookahead, that can further improve faithfulness upon the base decoding methods. Finally, we show a simple (and optionally iterative) distillation trick where the training of a student model incorporates the summaries generated with more faithfulness-aware methods, and the student model generates summaries of similar faithfulness with minimal decoding time.

Future experiments could extend similar analysis of faithfulness and factuality beyond summarization and develop a combination of heuristics that also encompasses other aspects and styles.

\section{Limitations}
While the decoding strategies with lookahead show improvement in faithfulness, they require a heavy computational overhead, especially when they are coupled with beam search for the base decoding strategy and for generating the future summary. We provide one solution with our distillation to improve decoding speed. Many of the computations, including the generated future summaries and the faithfulness scores on them, during this online process, are also later disregarded, similar to how any candidates are pruned during beam search. We believe an interesting direction might be to store the already generated future summaries so that the decoding may directly use the future summary if it is considered a good summary candidate.

\section{Ethical Impact}
While our work aims to reduce potential malicious or unintended harmful effects, our methods rely on the use of faithfulness metrics. The inherent problems and biases when using such metrics have been under-studied. Our decoding strategies can also be applied to be used for other metrics, even those that could be optimized for malicious intents. Another aspect to consider is the environmental impact of our proposed methods, as they require large computations. We hope that our distillation can mitigate this problem and future work can work towards more environmentally friendly approaches while improving faithfulness for safer use of large models.

\bibliography{anthology,custom}
\bibliographystyle{acl_natbib}

\appendix

\begin{figure*}[!t]
    \centering
    \includegraphics[width=\textwidth]{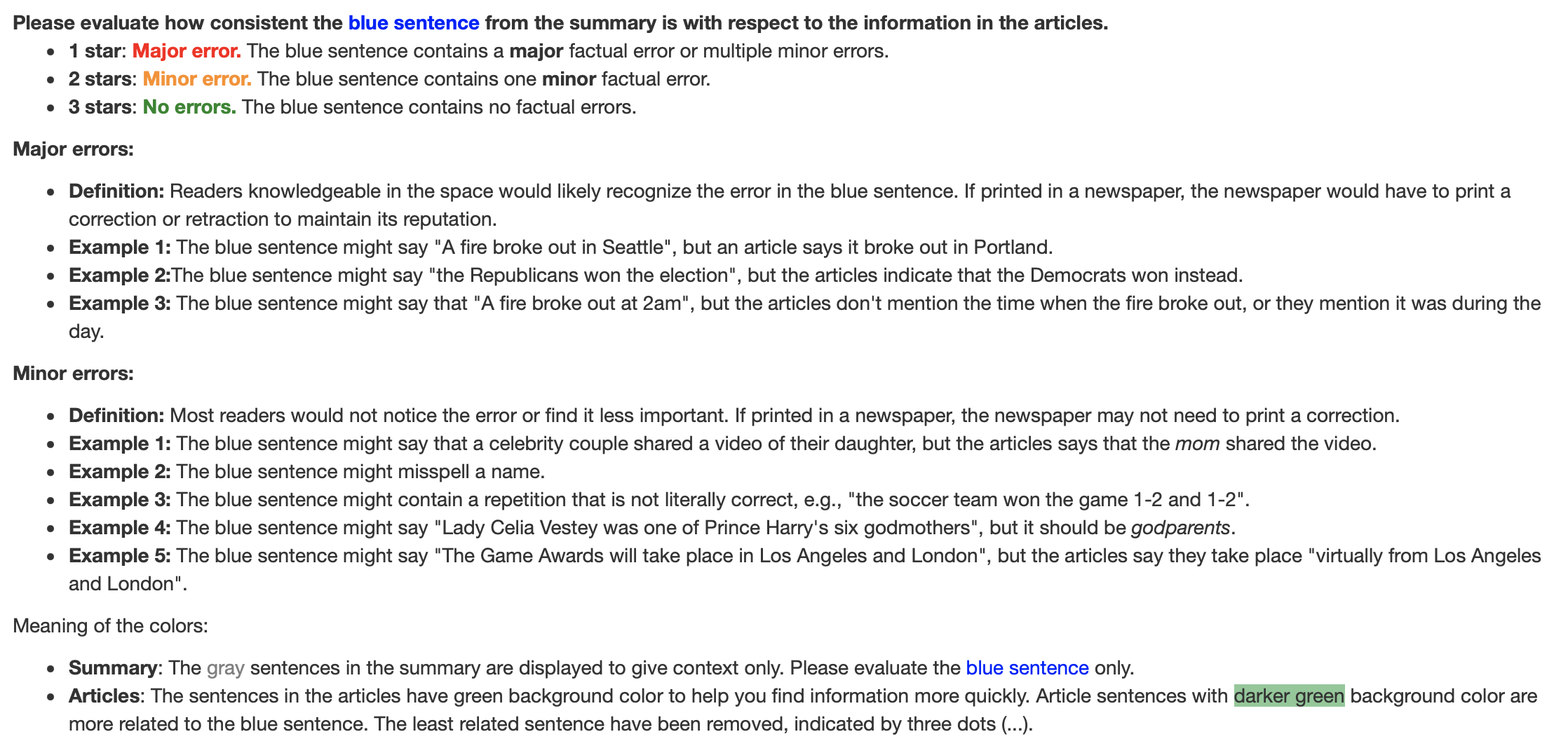}
    \caption{Annotation instructions to annotate factual consistency on Mechanical Turk.
    }
    \label{fig:annotationinstructions}
\end{figure*}

\begin{figure*}
    \centering
    \includegraphics[width=\textwidth]{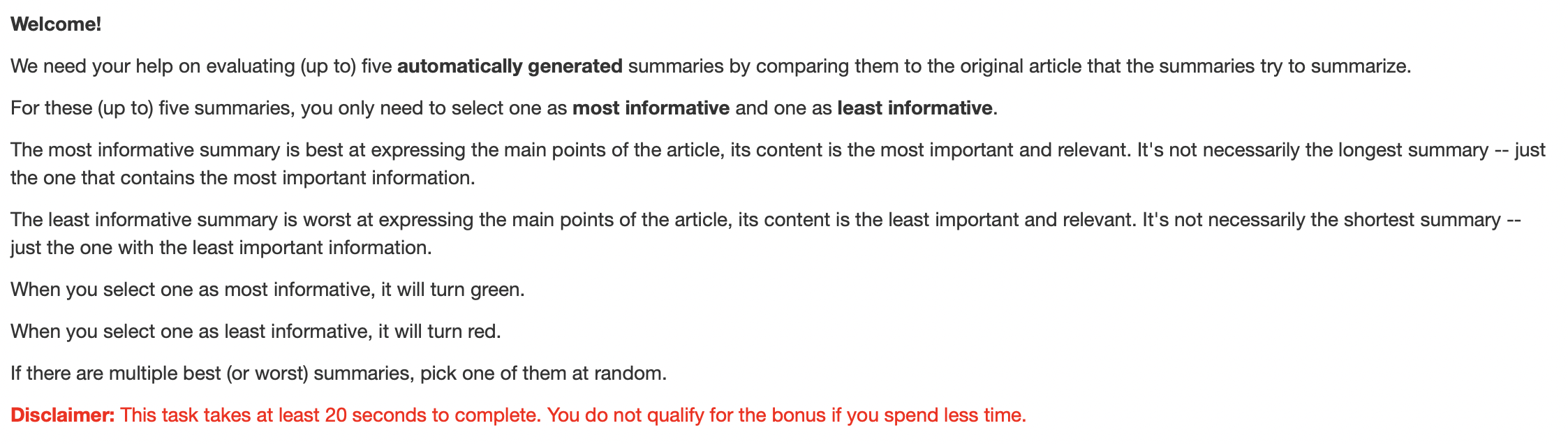}
    \caption{Annotation instructions to annotate informativeness on Mechanical Turk.}
    \label{fig:human_annotation_informativenss}
\end{figure*}

\section{Experiments Details.}\label{sec:experiment_detail_appendix}

\subsection{Datasets}\label{sec:dataset_appendix}
We evaluate on XSum and CNN/DM. We use the dataset processed and provided by \textsc{datasets} \cite{lhoest-etal-2021-datasets}.\footnote{The link to the processed data is found in \url{https://huggingface.co/datasets/xsum} and \url{https://huggingface.co/datasets/cnn_dailymail}.} Both datasets contain English news articles and the corresponding summaries. XSum contains 204045, 11332, and 11334 examples for training, validation, and test set, respectively, and CNN/DM contains 287113, 13368, and 11490 for the splits.

\subsection{Metrics}\label{sec:metrics_appendix}
We use the official code and follow the instructions to set up and run all the metrics we used. We use the ROUGE package from \url{https://github.com/google-research/google-research/tree/master/rouge}. We report all scores of our models from single runs. For BS and BS-Fact, we use the default model for English (\textsc{roberta-large}). For DAE, we use the sentence error, which considers the sentence to contain an error if one of its arcs is predicted to be not factual.

\subsection{Decoding Details}\label{sec:decoding_details_appendix}
\paragraph{Basic Decoding Method Details.}\label{sec:basic_appendix}
We use the official generation code provided by \textsc{transformers} \cite{wolf-etal-2020-transformers}. We use a single NVIDIA V100 GPU to generate the summaries. Greedy and sampling experiments take around 2 hours and beam search variants take 4 hours.

\paragraph{Ranking.} We do not need to do additional computation as we already have the outputs of beam search and the metric scores.

\paragraph{Lookahead.} To reduce computational overhead, we only calculate and incorporate the lookahead heuristics for the top 5 tokens according to the model score at each time step. Experiments using beam search or sampling to generate a future summary can be found in Appendix~\ref{sec:lookahead_appendix}.
We show additional ablations on the length of the future summaries and how the exploration changes with the heuristics in Appendix~\ref{sec:ablations_appendix}.
For tuning the $w$, the weight for the heuristics, we search over the interval from 5 to 55 with a step of 5, and evaluate the generated summaries on the development set. We use the average of all metric scores, including RL and BS so that we do not over-optimize for faithfulness. We find 25 to be optimal for CNN/DM and 55 for XSum. The time to run for XSum is 33 hours, and that for CNN/DM is around 70 hours.

\paragraph{Distillation Details.}
We use the example code from \textsc{transformers} to train summarization models. We follow the authors' hyper-parameters to train BART-Large models. We use 8 V100 GPUs and the training time is around 5 hours. To generate the summaries for the training data, The BS-Fact Ranker takes around 3 hours to generate the summaries when parallelized across the 8 GPUs. Lookahead takes 10 hours to generate the training data split across 8 GPUs.

\subsection{Human Evaluation Details}\label{sec:human_evaluation_appendix}

\begin{table}[!t]
    \centering
    \resizebox{0.99\columnwidth}{!}{\begin{tabular}{l c c c c c c }
    \toprule
    & RL & BS & BS-Fact & FactCC & DAE $\downarrow$ & QuestEval \\
    \midrule
    \multicolumn{7}{c}{CNN/DM} \\
    \midrule
    Greedy & \textbf{30.93} & \textbf{88.39} & 93.15 & 69.61 & 8.15 & 59.13 \\
    Nucleus $p=0.1$ & 30.93 & 88.39 & 93.15 & 69.58 & 8.17 & 59.13 \\
    Nucleus $p=0.3$ & 30.77 & 88.37 & 93.12 & 69.43 & 8.40 & 59.01 \\
    Nucleus $p=0.5$ & 30.39 & 88.31 & 92.95 & 66.67  & 10.05 & 58.81 \\
    Nucleus $p=0.7$ & 29.43 & 88.20	& 92.56 & 60.88 & 12.64	& 58.11 \\
    Nucleus $p=0.9$ & 27.64 & 87.90 & 91.76 & 54.05 & 21.61 & 56.43 \\
    Beam $k=2$  & 30.78 & 88.29 & 93.62 & 76.00 & 5.80 & 59.72 \\
    Beam $k=4$ & 30.44 & 88.17 & 93.95 & 81.19 & 4.11 & 59.97 \\
    Beam $k=6$ & 30.30 & 88.12 & 94.07 & 82.94 & 3.76 & \textbf{60.07} \\
    Beam $k=8$ & 30.10 & 88.07 & 94.15 & 83.50 & 3.39 & 60.06\\
    Beam $k=10$ & 29.99 & 88.03 & \textbf{94.20} & \textbf{84.23} & \textbf{3.30} & 60.03 \\
    \midrule
    \multicolumn{7}{c}{XSum} \\
    \midrule
    Greedy  & 36.16 & 92.03 & 89.28 & \textbf{23.53} & 65.35 & 36.51 \\
    Nucleus $p=0.1$ & 31.08 & 91.24 & 88.61 & 21.63 & 76.07 & 35.04\\
    Nucleus $p=0.3$ & 31.19 & 91.25 & 88.62 & 21.51  & 76.35 & 34.98 \\
    Nucleus $p=0.5$ & 31.08 & 91.24 & 88.63 & 21.11 & 75.31 & 34.99 \\
    Nucleus $p=0.7$ & 31.24 & 91.24 & 88.62	& 21.26	& 76.11	& 35.01\\
    Nucleus $p=0.9$ & 31.15 & 91.26 & 88.62 & 21.04 & 76.20 & 34.98 \\
    Beam $k=2$ & 36.76 & 92.13 & 89.38 & 22.98 & 64.62 & 36.82 \\
    Beam $k=4$ & 36.96 & 92.14 & 89.42 & 23.00 & 63.81 & 36.97 \\
    Beam $k=6$ & 37.09 & \textbf{92.14} & 89.43 & 22.70 & 63.71 & 37.00 \\
    Beam $k=8$ & 37.09 & 92.13 & 89.44 & 23.05 & 63.52 & 37.02 \\
    Beam $k=10$ & \textbf{37.11} & 92.12 & \textbf{89.45} & 22.97 & \textbf{63.49} & \textbf{37.05}\\
    \bottomrule
    \end{tabular}
    }
    \caption{Full results of beam search and nucleus sampling for fine-tuned \textsc{Bart-Large} models. The trend can still be seen under different beam sizes and top-$p$ values, where increasing $k$ improves faithfulness and increasing $p$ degrades it.}
    \label{tab:analysis_full}
\end{table}

\paragraph{Human Evaluation on Faithfulness.} The screenshot of the annotation can be seen in \autoref{fig:annotationinstructions}. We required annotators to pass a custom qualification test consisting of three summaries with factual errors. To pass the test, the annotators had to correctly describe the factual errors in words. Workers also needed to have previously completed 100 or more tasks with an acceptance rate of 95\%
or higher. We recruited workers from countries whose main language is English.
 To prevent any one worker from dominating the results, we set a maximum of 100 HITs per worker per dataset. The payment for judging each summary was \$0.22 plus a bonus of \$0.03. Annotators who spent more than 10 seconds per HIT and maintained high accuracy on HITs with known answers obtained the bonus. Annotators spent a median amount of 57.5 seconds per HIT, which amounts to a pay of \$15.65 per hour. Krippendorff alpha \cite{krippendorff1980content} for the CNN/DM factuality annotation is 0.63, and Krippendorff alpha for the XSum annotation is 0.57.

\paragraph{Human Evaluation on Informativeness.} The screenshot of the annotation can be seen in \autoref{fig:human_annotation_informativenss}. To achieve good quality, we set up a qualification task of three documents with their associated summaries. A selected pool of workers who had passed previous factuality qualification tests was allowed to take this current qualification test. The workers who passed the current qualification test were allowed to participate in this evaluation. In addition, we added the same three documents with known answers to the evaluation and observed that workers had 100\% accuracy on them. We set the same maximum of 100 HITs per worker per dataset as in the factuality evaluation. The pay was \$0.40 plus \$0.10 bonus per HIT. Annotators spent a median time of 112 seconds per HIT, amounting to a pay of \$16.07 per hour.
For inter-annotator agreement, Krippendorff alpha \cite{krippendorff1980content} for the CNN/DM annotation is 0.22, and Krippendorff alpha for the XSum annotation is 0.32.

\begin{table}[!t]
    \centering
    \resizebox{0.99\columnwidth}{!}{\begin{tabular}{l c c c c c c}
    \toprule
     & RL & BS & BS-Fact & FactCC & DAE $\downarrow$ & QuestEval \\
    \midrule
    \multicolumn{7}{c}{CNN/DM} \\
    \midrule
    Greedy & 30.20 & 87.65 & 89.71 & 53.00 & 15.44 & 56.70 \\
    Nucleus $p=0.1$  & 30.20 & 87.65 & 89.71 & 52.99 & 15.44 & 56.68\\
    Nucleus $p=0.3$  & 30.15 & 87.64 & 89.71 & 52.96 & 15.72 & 56.68 \\
    Nucleus $p=0.5$  & 29.88 & 87.61 & 89.62 & 51.48 & 17.25 & 56.43 \\
    Nucleus $p=0.7$  & 28.86 & 87.47 & 89.36 & 46.44 & 20.97 & 55.92\\
    Nucleus $p=0.9$  & 30.15 & 87.23 & 88.88 & 38.82 & 28.53 & 54.78 \\
    Beam $k=2$ & 30.67  & \textbf{87.72}  & 90.28  & 57.75  & 11.40  & 57.24 \\
    Beam $k=4$ & \textbf{30.82}  & 87.71  & 90.61  & 62.36  & 9.50  & 57.42 \\
    Beam $k=6$ & 30.67  & 87.66  & 90.75  & 64.18  & 8.72  & 57.43\\
    Beam $k=8$ & 30.68  & 87.65  & 90.82  & 64.96  & 8.66  & 57.43 \\
    Beam $k=10$ & 30.66 & 87.65  & \textbf{90.87}  & \textbf{65.32}  & \textbf{8.23}  & \textbf{57.46} \\
    \midrule
    \multicolumn{7}{c}{XSum} \\
    \midrule
    Greedy & 38.53 & 92.45 & 89.05 & \textbf{24.53} & 68.33 & 35.75 \\
    Nucleus $p=0.1$ & 38.53 & 92.44 & 89.05 & 24.50 & 68.33 & 35.76\\
    Nucleus $p=0.3$ & 38.42 & 92.42 & 89.02 & 24.10	 & 69.23 & 35.68\\
    Nucleus $p=0.5$ & 37.85 & 92.33 & 88.97 & 23.14 & 70.07 & 35.51\\
    Nucleus $p=0.7$ & 36.13 & 92.09 & 88.80 & 22.95	 & 72.72 & 35.27\\
    Nucleus $p=0.9$ & 33.76 & 91.68 & 88.51 & 22.46 & 76.23 & 34.73\\
    Beam $k=2$ & 39.09 & 92.53 & 89.13 & 23.58 & 67.72 & 35.90 \\
    Beam $k=4$  & 39.35 & \textbf{92.58} & 89.19 & 22.64 & 67.16 & 35.97 \\
    Beam $k=6$  & 39.32 & 92.57 & 89.21 & 22.89 & 66.86	& \textbf{35.98} \\
    Beam $k=8$  & 39.37 & 92.57 & 89.21 & 22.71 & 66.73 & 35.97\\
    Beam $k=10$ & \textbf{39.43} & 92.57 & \textbf{89.23} & 22.75 & \textbf{66.48} & 35.96 \\
    \bottomrule
    \end{tabular}
    }
    \caption{Full results of beam search and nucleus sampling for fine-tuned \textsc{PEGASUS-Large} models. We observe a similar observation as \autoref{tab:analysis_full}, showing that the faithfulness trend holds for different models.}
    \label{tab:anlysis_full_pegasus}
\end{table}

\begin{table*}[!t]
    \centering
    \small
    \begin{tabular}{l c c c c c c}
    \toprule
     & Rouge-L & BERTScore & BS-Fact & FactCC & DAE $\downarrow$  & QuestEval \\
    \midrule
    \multicolumn{7}{c}{CNN/DM} \\
    \midrule
    Greedy &  \textbf{30.93} & \textbf{88.39} & 93.15 & 69.61 & 8.15 & 53.71 \\
    Beam $k=10$ & 29.99 & 88.03 & 94.20 & 84.23 & 3.30 & 60.03 \\
    Greedy + Greedy Lookahead  & 30.88 & 88.38 & 93.57 & 71.54 & 6.42 & 59.70 \\
    Greedy + Sampling Lookahead  & 30.67 & 88.35 & 93.54 & 78.28 & 7.09  & 59.72 \\
    Greedy + Beam Lookahead  & 30.63 & 88.32  & 93.85 & 82.07 & 5.33	& 60.13 \\
    Beam + Greedy Lookahead  & 28.66 & 87.84 & \textbf{95.32} & \textbf{86.10} & \textbf{1.68} & \textbf{63.69} \\
    \midrule
    \multicolumn{7}{c}{XSum} \\
    \midrule
    Greedy & 36.16 & 92.03 & 89.28 & 23.53 & 65.35 & 36.51 \\
    Beam $k=10$ & \textbf{37.11} & \textbf{92.12} & 89.45 & 22.97 & 63.49 & 37.05 \\
    Greedy + Greedy Lookahead & 36.25 & 92.11 & 89.71 & \textbf{24.21} & 60.46 & 37.17 \\
    Greedy + Sampling Lookahead & 36.24 & 92.10 & 89.55 & 23.97 & 62.35 & 36.90 \\
    Greedy + Beam Lookahead & 36.17 & 92.07 & 89.62 & 23.58 & 61.90 & 37.10 \\
    Beam + Greedy Lookahead & 35.27 & 91.94 & \textbf{90.78} & 23.38 & \textbf{50.04} & \textbf{39.24} \\
    \bottomrule
    \end{tabular}
    \caption{Lookahead results with different decoding strategies for base decoding strategies and the lookahead generation strategies.}
    \label{tab:lookahead}
\end{table*}

\section{Full Analysis}\label{sec:analysis_appendix}
\autoref{tab:analysis_full} shows the full result. We see the general trend where increasing beam size improves faithfulness and increasing $p$ for sampling is not helpful for faithfulness.

We similarly run the experiment on PEGASUS, a 568M model specifically trained for the task of abstractive summarization, with its respective checkpoints.\footnote{We use the checkpoint from \url{https://huggingface.co/google/pegasus-cnn_dailymail} and \url{https://huggingface.co/google/pegasus-xsum}.} The result is presented in \autoref{tab:anlysis_full_pegasus}.

\begin{table}[!t]
    \centering
    \resizebox{0.99\columnwidth}{!}{\begin{tabular}{c c c c c c c c c c c c c c}
    \toprule
    & \multicolumn{4}{c}{All} & \multicolumn{4}{c}{CNN/DM} & \multicolumn{4}{c}{XSum} \\
    & \multicolumn{2}{c}{Pearson} & \multicolumn{2}{c}{Spearman} & \multicolumn{2}{c}{Pearson} & \multicolumn{2}{c}{Spearman} & \multicolumn{2}{c}{Pearson} & \multicolumn{2}{c}{Spearman} \\
    & $\rho$ & $p$ & $r$ & $p$ & $\rho$ & $p$ & $r$ & $p$ & $\rho$ & $p$ & $r$ & $p$ \\
    \midrule
    FactCC* & .20 & .00 & .30 & .00 & .36 & .00 & .30 & .00 & .07 & .07 & .19  & .00 \\
    DAE* & .18 & .00 & .20 & .00 & .27 & .00 & .22 & .00 & .03 & .38 & \textbf{.33} & .00 \\
    BS-Fact* & .30 & .00 & .25 & .00 & .38 & .00 & .31 & .00 & .20 & .00  & .09 & .02 \\
    QuestEval & .19 & .00 & .20 & .00 & .21 & .00 & .19 & .00 & .16 & .00 & .09 & .00 \\
    \midrule
    Comp. Avg & .34 & .00 & .32 & .00 & .30 & .00 & .33 & .00 & .30 & .00 & .32 & .00  \\
    Comp. Tuned & \textbf{.37} & .00 & \textbf{.34} & .00 & \textbf{.42} & .00 & \textbf{.36} & .00 & \textbf{.31} & .00 & .19 & .00 \\
    \bottomrule
    \end{tabular}
    }
    \caption{Partial correlations of metrics on the Frank test dataset. Composite achieves the highest correlations on the combined and XSum dataset. * indicates results copied from the original work.}
    \label{tab:composite_correlation}
\end{table}

\section{Lookahead Methods} \label{sec:lookahead_appendix}
We show the result of combining different decoding strategies for the base decoding strategy as well as for lookahead in \autoref{tab:lookahead} shows the result. We experiment with greedy and beam search as the base decoding strategies. For greedy, we experiment with all three decoding strategies for lookahead. For beam search, we are unable to run it with sampling or beam search due to the large computational cost. Interestingly, using beam for lookahead does not provide additional gains. We suspect that this is because exploring the future with more beams cannot guarantee that the base decoding strategy is able to explore them, as it is limited to selecting only the top tokens.

\section{Composite Metric} \label{sec:composite_metric_appendix}

As described in Section~\ref{sec:ranking}, we train the composite metric on \textsc{FactCollect} and tune it on FRANK \cite{pagnoni-etal-2021-understanding}. We use the test set of \citet{pagnoni-etal-2021-understanding} for evaluation and the rest for tuning the composite metric.
The resulting weights for the metrics are 0.29, -0.29, 1.97, and 0.94 for the FactCC, DAE, BS-Fact, and QuestEval, respectively, and the intercept is -1.91.
We additionally compute partial correlations on FRANK, shown in \autoref{tab:composite_correlation}. We see that the composite is able to further increase the correlations in all settings except for XSum's Spearman correlation. Ablations on the effect of ranking with a single metric in Appendix~\ref{sec:ablations_appendix}.

Since \textsc{FactCollect} only contains annotations on XSum and CNN/DM, we analyze whether the composite metric is robust for another dataset and domain. We use WikiHow \cite{koupaee2018wikihow} and decode using PEGASUS\footnote{We use the checkpoint \textsc{pegasus-wikihow} (\url{https://huggingface.co/google/pegasus-wikihow}).} with greedy and beam decoding. The result of applying ranking to the beam output can be seen in \autoref{tab:ranking_wikihow}. We see consistent gains in all faithfulness metrics when we apply ranking, showing its robustness of improving faithfulness in another domain.

\begin{table}[!t]
    \centering
    \resizebox{0.99\columnwidth}{!}{\begin{tabular}{l c c c c c c}
    \toprule
     & RL & BS & BS-Fact & FactCC & DAE $\downarrow$  & QuestEval \\
     \midrule
     Greedy & 26.36 & \textbf{89.09} & 88.93 & 89.57 & 75.34 & 38.39 \\
     Beam & 27.52 & 87.56 & 89.41 & 87.20 & 60.21 & 39.44 \\
     \ranking{} & \textbf{27.60} & 87.62 & \textbf{89.64} & \textbf{91.11} & \textbf{47.01} & \textbf{41.91} \\
     \bottomrule
    \end{tabular}
    }
    \caption{Results for ranking on the WikiHow dataset.}
    \label{tab:ranking_wikihow}
\end{table}

\begin{figure*}[!t]
    \centering
    \begin{subfigure}{.4\textwidth}
    \includegraphics[width=\textwidth]{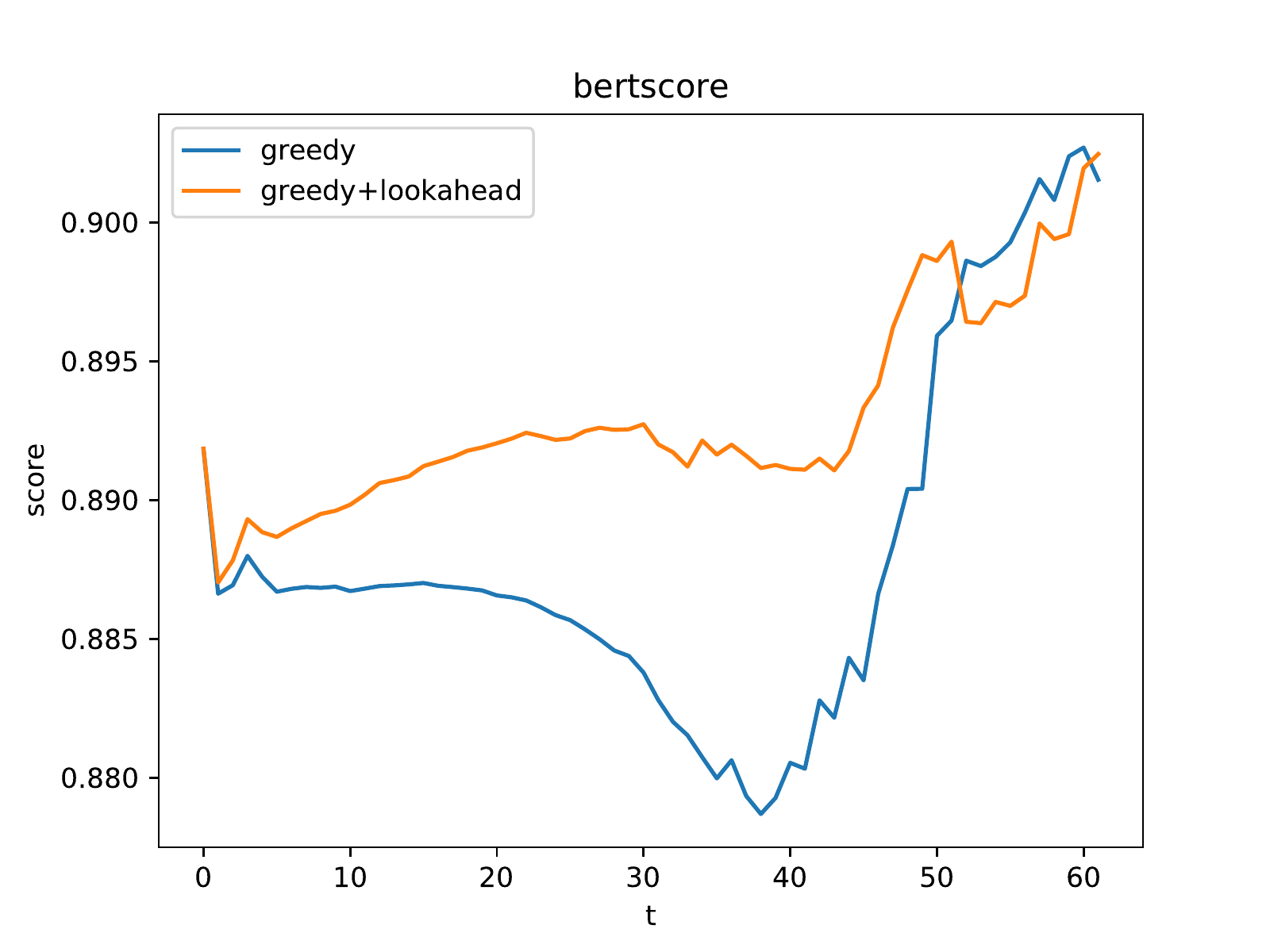}
    \end{subfigure}
    \begin{subfigure}{.4\textwidth}
    \includegraphics[width=\textwidth]{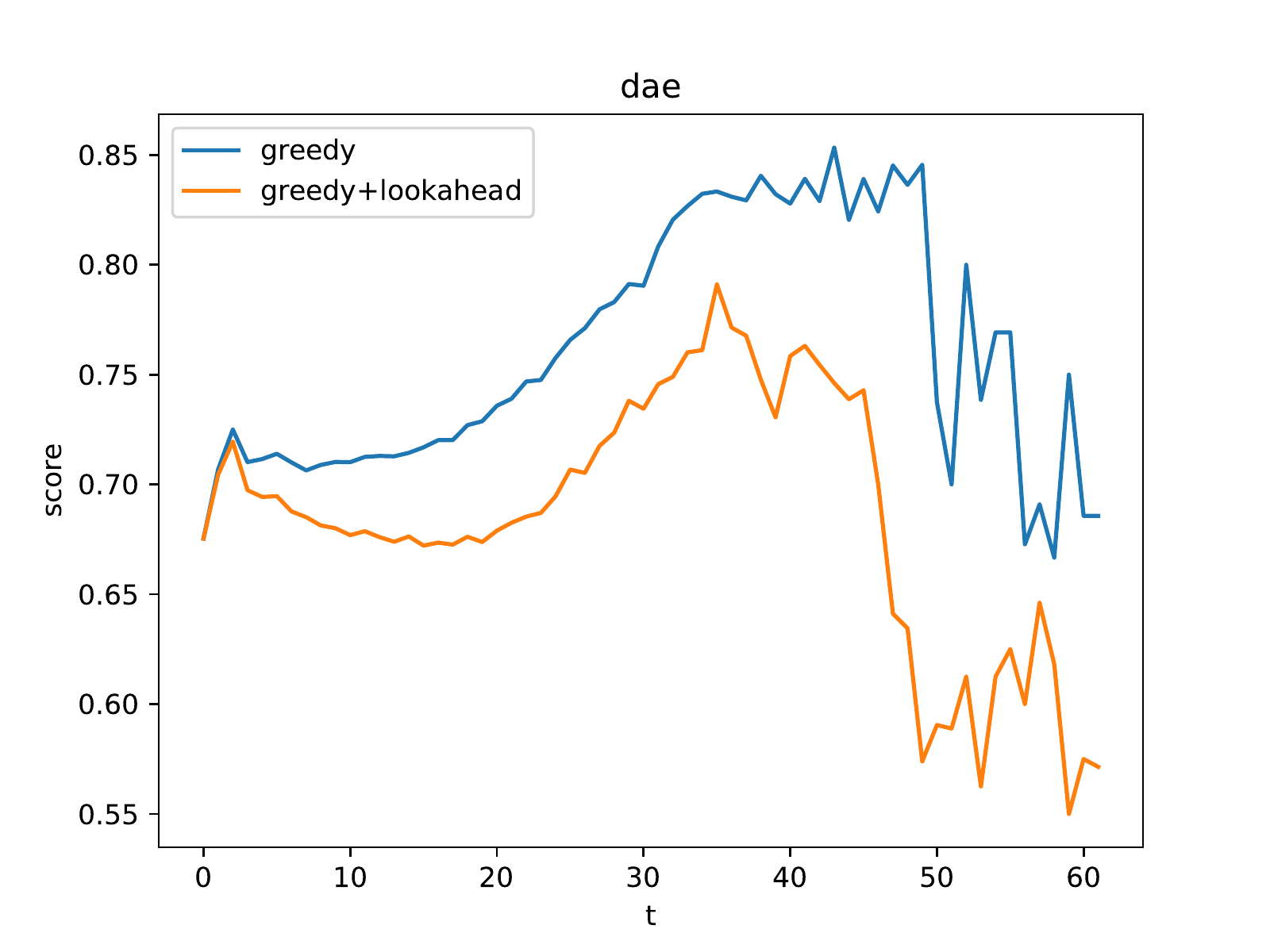}
    \end{subfigure}
    \caption{Faithfulness score of the lookahead summaries at each time step. Adding lookahead as the heuristics improves the search space to generate more faithful summaries.}
    \label{fig:ablation}
\end{figure*}

\section{Ablations} \label{sec:ablations_appendix}
\begin{table}[t!]
    \centering
    \resizebox{0.99\columnwidth}{!}{\begin{tabular}{l c c c c c c}
    \toprule
    & RL & BS & BS-Fact & FactCC & DAE $\downarrow$  & QuestEval \\
    \midrule
    \multicolumn{7}{c}{CNN/DM} \\
    \midrule
    Greedy & \textbf{30.93} & \textbf{88.39} & 93.15 & 69.61 & 8.15 & 59.13 \\
    $l=0$ & 30.71 & 88.34 & 93.13 & 70.41 & 8.19 & 58.65 \\
    $l=$ full & 30.75 & 88.35 & \textbf{93.90} & \textbf{71.54} & \textbf{5.70} & \textbf{60.13} \\
    \midrule
    \multicolumn{7}{c}{XSum} \\
    \midrule
    Greedy & 36.16 & 92.03 & 89.28 & 23.53 & 65.35 & 36.51 \\
    $l=0$  & 35.73 & 92.00 & 89.39 & 23.43 & 64.06 & 36.55 \\
    $l=$ full & \textbf{36.25} & \textbf{92.11} & \textbf{89.71} & \textbf{24.21} & \textbf{60.46} & \textbf{37.17} \\
    \bottomrule
    \end{tabular}
    }
    \caption{Lookahead ablation with different lengths. $l=0$ provides the faithfulness heuristic score only on the partially generated summaries while $l=\text{full}$ is our lookahead model that evaluates on the full future summary. The faithfulness score calculated on the partial summaries does not provide an effective estimate that improves the faithfulness of the generated summary.
    }
    \label{tab:ablation_length}
\end{table}

We present several ablation studies for our proposed faithfulness-aware decoding methods. More ablation studies exploring how lookahead explores the search space can be found in Appendix~\ref{sec:ablations_appendix}.

\paragraph{Lookahead Length.} We first present the result of using the lookahead heuristics but with $l=0$. This means that at each time step, we do not use future heuristics but directly evaluate the faithfulness of the already generated partial summaries as the additional score. The result using \textsc{Greedy+Lookahead} is shown in \autoref{tab:ablation_length}. Compared to greedy decoding, adding the faithfulness score of the current partial summary shows mixed results; the heuristic can only slightly improve BS-Fact, DAE, and QuestEval for XSum. However, we only see substantial gain when the future is taken into account (i.e. $l=\text{full}$). This shows the necessity of using the full summary to achieve the full potential of current faithfulness metrics.

\paragraph{Ranking with Faithfulness Metrics.} Next, we present the result for ranking with each respective faithfulness metric. The result is shown in \autoref{tab:ranking_experiment}.
Generally, optimizing for one metric will lead to improvement in other faithfulness metrics.
While optimizing each of the faithfulness metrics will undoubtedly perform the best when we use that metric for evaluation, the composite metric is able to achieve a similarly good score for \textbf{all} faithfulness metric that we are considering.

\begin{table}[t!]
    \centering
    \resizebox{0.99\columnwidth}{!}{\begin{tabular}{l c c c c c c c c c}
    \toprule
    Ranker & RL & BS & BS-Fact & FC & DAE $\downarrow$  & QE & COMP \\
    \midrule
    \multicolumn{7}{c}{CNN/DM} \\
    \midrule
    First & 29.99 & 88.03 & 94.20 & 84.23 & 3.30 & 60.03 & 74.68 \\
    BS-Fact & 29.81 & 88.04 & \textbf{94.64} & 84.08 & 3.04 & 60.41 &  75.94 \\
    FC & 29.98 & 88.04 & 94.20 & \textbf{90.75} & 3.00 & 60.06 & 76.75 \\
    DAE & 30.00 & 88.03 & 94.20 & 84.28 & \textbf{1.92} & 60.04 & 75.11 \\
    QE & \textbf{30.27} & \textbf{88.16} & 94.14 & 82.81 & 2.83 & \textbf{63.26} & 77.33 \\
    Comp. & 30.08 & 88.12 & 94.31 & 90.27 & \textbf{1.92} & 62.57 & \textbf{79.51} \\
    \midrule
    \multicolumn{7}{c}{XSum} \\
    \midrule
    Top & \textbf{37.11} & 92.12 & 89.45 & 22.97 & 63.49 & 37.05 & 8.08 \\
    BS-Fact & 36.46 & \textbf{92.14} & \textbf{90.15} & 22.10 & 56.33 & 37.98 & 12.15 \\
    FC & 36.98 & 92.11 & 89.44 & \textbf{41.93} & 63.47 & 37.01 & 13.67 \\
    DAE & 36.94 & 92.11 & 89.54 & 23.27 & \textbf{50.82} & 37.28 & 12.24 \\
    QE & 36.36 & 92.06 & 89.61 & 23.07 & 60.35 & \textbf{41.17} & 13.20 \\
    Comp. & 36.42 & 92.10 & 89.79 & 40.11 & 51.48 & 40.10 & \textbf{20.20} \\
    \bottomrule
    \end{tabular}
    }
    \caption{Ranking results with different faithfulness metrics. Top is the best summary from beam search, and each subsequent rows represent the ranker using the corresponding faithfulness metric. We abbreviate FactCC as FC, QuestEval as QE, and Composite as Comp.
    }
    \label{tab:ranking_experiment}
\end{table}

\paragraph{Evaluating the Search Space.} We hypothesize that by incorporating lookahead, we can improve the search space even when a few tokens are generated. To better understand this, we greedily decode the full summary at each time step given the prefix similar to how  lookahead works. We then use BS-Fact and DAE to score all generated summaries and analyze the faithfulness score at each time step. Here, we focus on XSum and compare greedy and \textsc{Greedy+Lookahead}. The plots of faithfulness scores using the current prefix to generate the full summaries are shown in \autoref{fig:ablation}, where we see the benefit of having the lookahead heuristics.
For BS-Fact, we see a large gap between the two methods especially when t is between 5 and 50. Though it may be less surprising as this is the faithfulness metric that the lookahead heuristic optimizes on, the heuristic can nevertheless prevent the score to dip, which we see for greedy search between $t=5$ to $t=40$. This shows that it is able to lead the model to a more faithful path to prevent straying away from a less faithful path. When we evaluate DAE, we show that optimizing on BS-Fact with lookahead heuristic can  consistently improve the score for all lengths.

\begin{figure}[!t]
    \centering
    \includegraphics[width=\columnwidth]{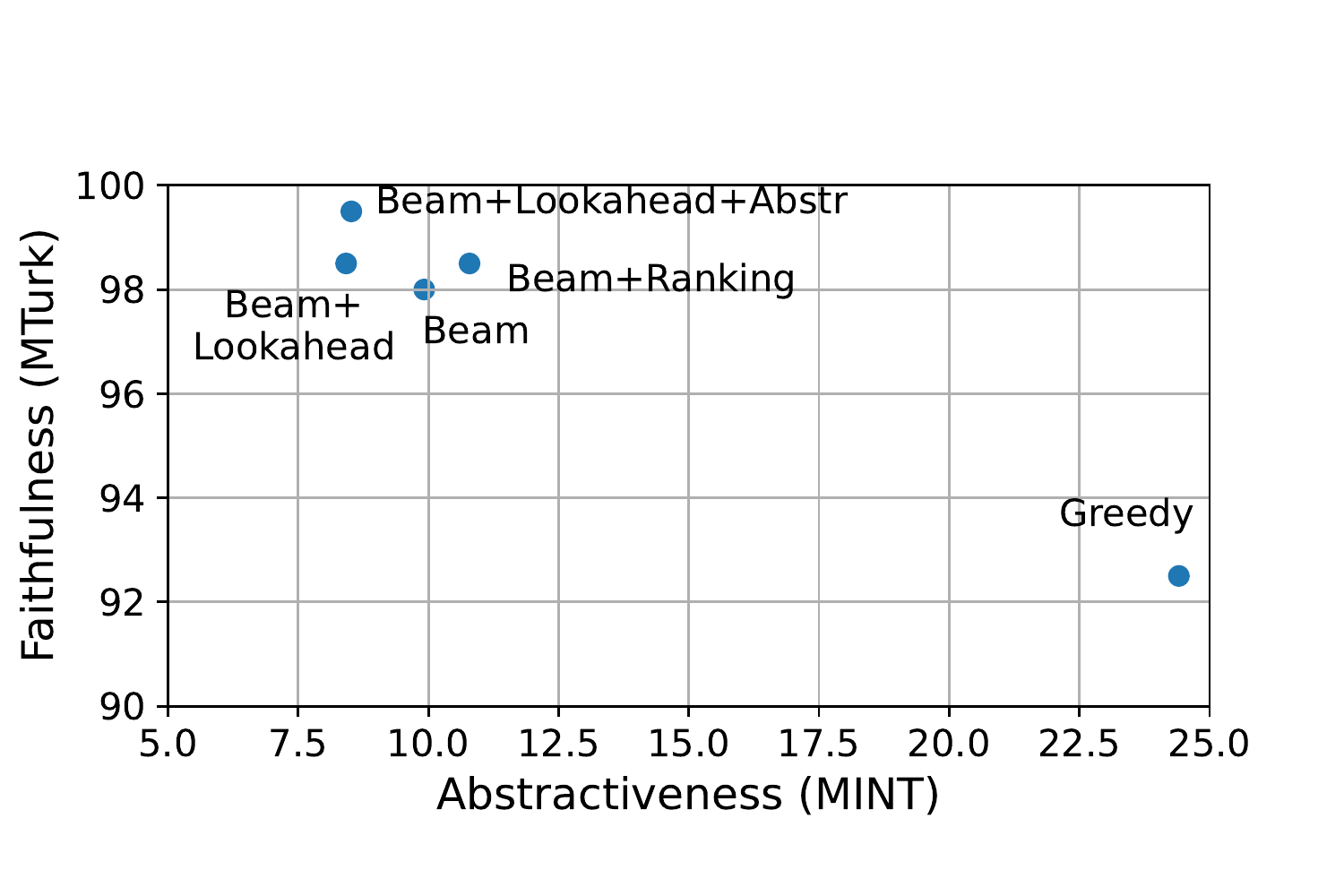}
    \caption{Faithfulness and abstractiveness tradeoff results on the 200 CNN/DM examples used for human annotation. While our proposed methods are less abstractive, the gain in faithfulness is much larger than the  decrease in abstractiveness.
    }
    \label{fig:abstractiveness_cnn}
\end{figure}

\begin{figure}[!t]
    \centering
    \includegraphics[width=0.95\columnwidth]{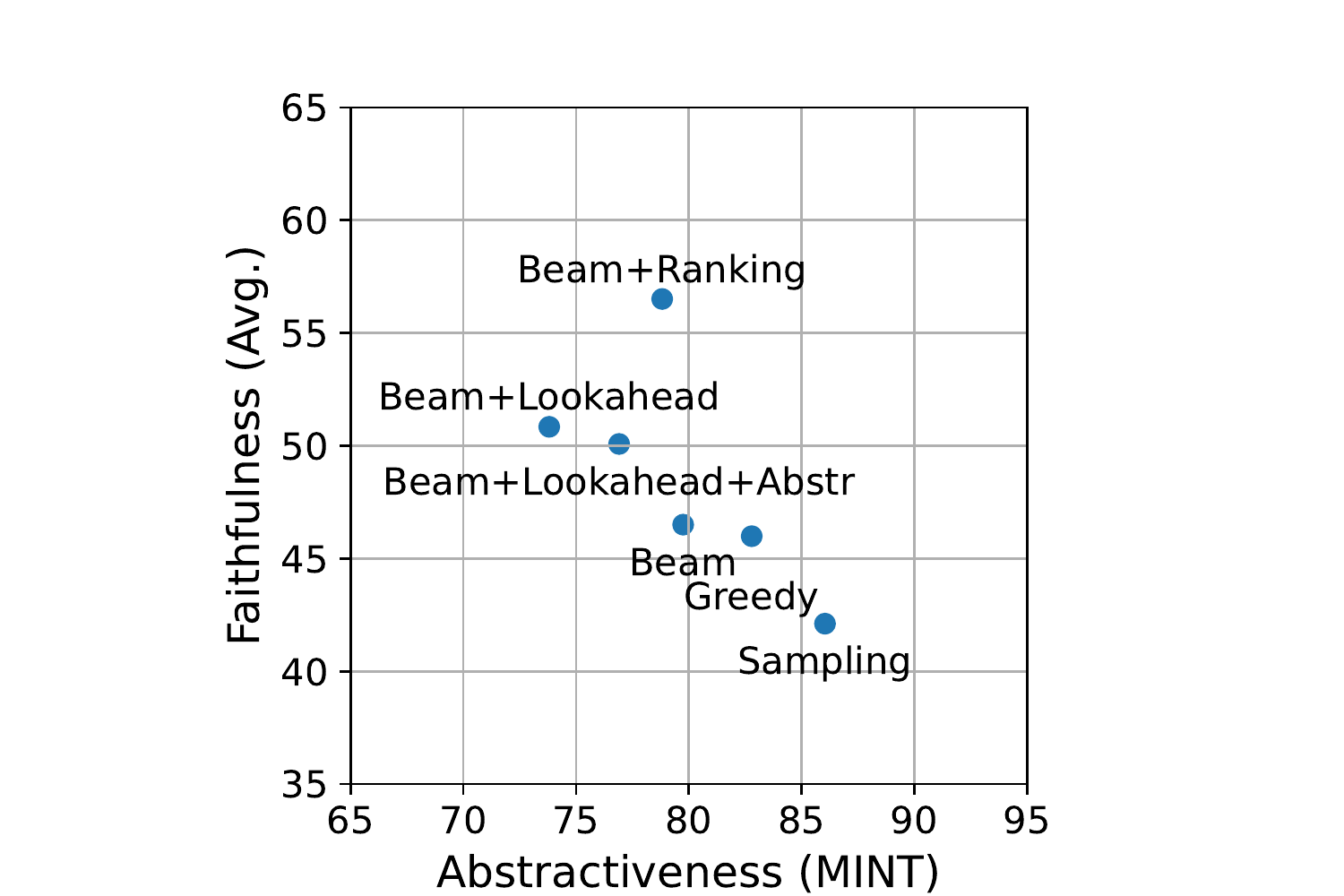}
    \caption{Faithfulness and Abstractiveness tradeoff results on the full examples XSum test set. Faithfulness is calculated by taking the average across all automatic faithfulness metrics.
    }
    \label{fig:faithfulness_abstractive_metric}
\end{figure}

\begin{figure}[!t]
    \centering
    \includegraphics[width=\columnwidth]{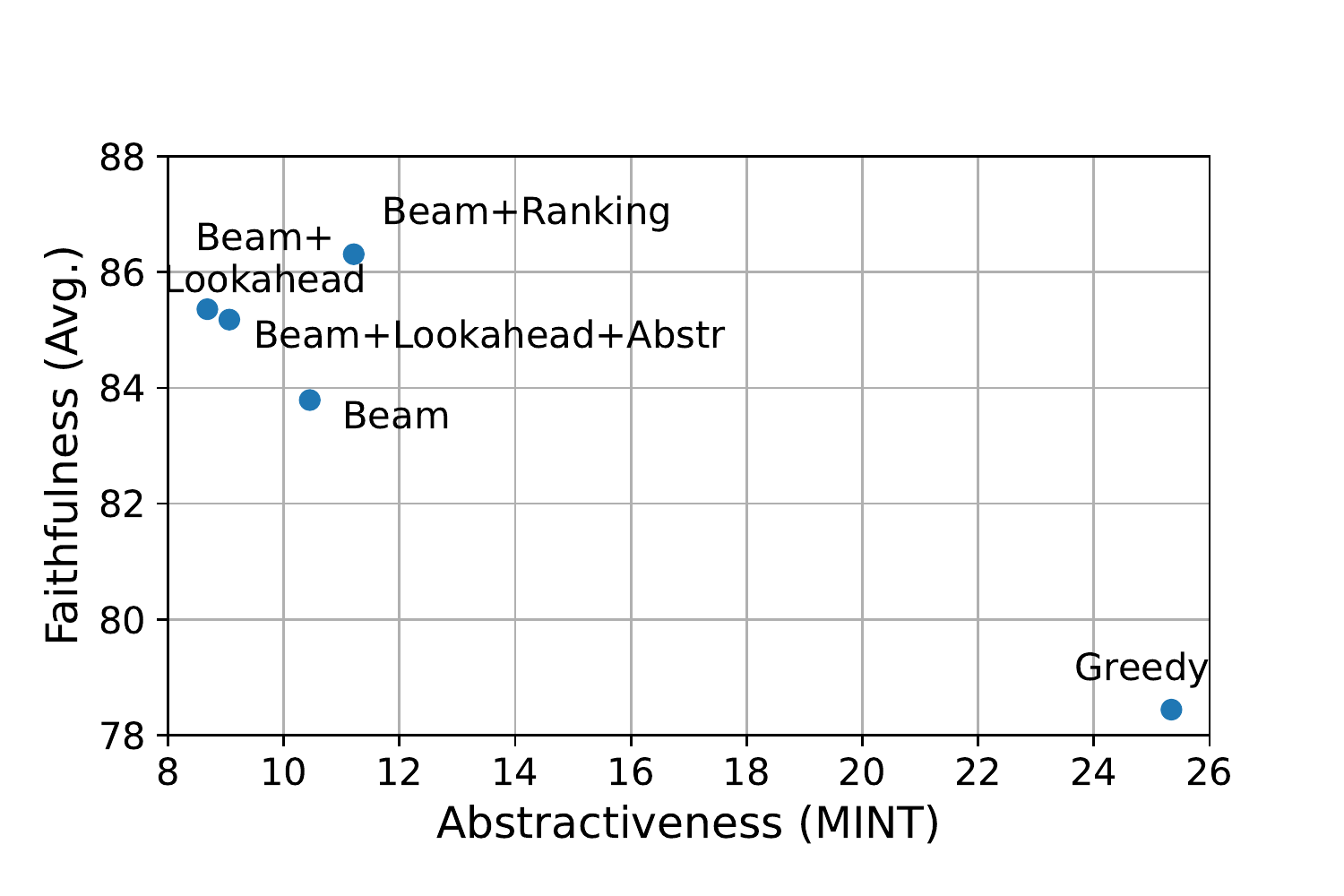}
    \caption{Faithfulness and Abstractiveness tradeoff results on the full examples CNN/DM test set. Faithfulness is calculated by taking the average across all automatic faithfulness metrics.
    }
    \label{fig:faithfulness_abstractive_metric_cnn}
\end{figure}

\section{Abstractiveness}\label{sec:abstractiveness_appendix}
We first show the same tradeoff result in CNN/DM in \autoref{fig:abstractiveness_cnn}. \textsc{Beam+Lookahead+Abstr} does achieve a slightly higher \textsc{Mint} score while also improving faithfulness. 

We also extend the analysis to the whole test dataset and show the faithfulness score by taking the average of all faithfulness metrics (Avg.). Since DAE is an error rate, we subtract the score from 100 so that a higher score means it is more faithful. We do not use the composite metric as the ranking directly optimizes for it.

We can see a similar trend with the average of faithfulness metrics for both datasets in \autoref{fig:faithfulness_abstractive_metric} and \autoref{fig:faithfulness_abstractive_metric_cnn}, where the gain in faithfulness outweighs the decrease in abstractiveness. The difference from the result using human faithfulness score is that \ranking{} achieves the highest average score since ranking with composite metric optimizes the faithfulness metrics.

\end{document}